%% file: main.tex
\documentclass[letterpaper]{article} 
\usepackage{aaai2026}  
\usepackage{times}  
\usepackage{helvet}  
\usepackage{courier}  
\usepackage[hyphens]{url}  
\usepackage{graphicx} 
\usepackage{natbib}  
\usepackage{caption} 
\usepackage{algorithm}
\usepackage{algorithmic}

\usepackage{booktabs}
\usepackage{multirow}
\usepackage{amsmath,amssymb}
\usepackage{dsfont}

\usepackage{newfloat}
\usepackage{listings}
\DeclareCaptionStyle{ruled}{labelfont=normalfont,labelsep=colon,strut=off} 
\floatstyle{ruled}
\newfloat{listing}{tb}{lst}{}
\floatname{listing}{Listing}
\title{Rethinking Efficient Crack Segmentation with Task-Aligned Structural-Directional Modeling}
\author{
    Shipeng Liu\textsuperscript{\rm 1},
    Liang Zhao\textsuperscript{\rm 1},
    Dengfeng Chen\textsuperscript{\rm 1},
    Weihua Zhang\textsuperscript{\rm 1}
}
\affiliations{
    \textsuperscript{\rm 1}Xi'an University of Architecture and Technology\\
    \{lsp, zhaoliang, chdfeng\}@xauat.com, weihuazhang0826@gmail.com

}

\usepackage{bibentry}

\begin{document}

\maketitle

\begin{abstract}
Recent crack segmentation methods often follow generic semantic segmentation designs, using stronger backbones, hybrid CNN-Transformer-Mamba encoders, and auxiliary enhancement branches. Although effective, this raises whether stronger generic feature mixing is the most suitable direction for crack segmentation. We instead formulate crack segmentation as sparse structural recovery. Cracks have limited category-level semantics but strong morphological regularities, being thin, sparse, anisotropic, locally fragmented, and easily confused with textures or shadows. Thus, the key bottleneck lies in preserving weak structural evidence, recovering directional continuity, and suppressing background coupling. We propose \textbf{RIFT}, a compact family of morphology-aligned crack segmentation models. Rather than compressing a complex generic architecture, RIFT is simple by design, preserving local evidence, aggregating cooperative directional continuity, and restoring crack structures through lightweight multi-scale fusion. Experiments on four public benchmarks show that RIFT achieves the best or tied-best results across the 16 main metrics against reproduced representative baselines. RIFT-B gives the strongest overall accuracy, while RIFT-T provides the best deployment efficiency with only 0.47M parameters and high inference speed. Topology-aware evaluation, ablations, transfer experiments, and visualizations further verify that task-aligned simplicity can match or surpass complex hybrid architectures when its inductive bias fits crack morphology. 
\end{abstract}

\textbf{Code}: \url{https://github.com/xauat-liushipeng/RIFT}

\section{Introduction}\label{sec1}

\begin{figure}[t]
    \centering
    \includegraphics[width=1\linewidth]{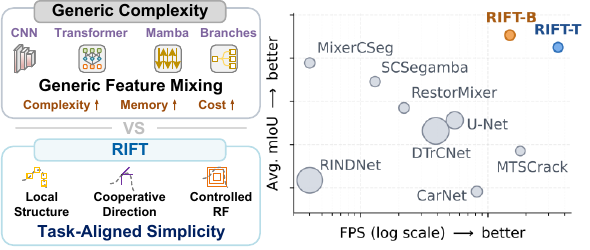}
    \caption{\textbf{Task-aligned simplicity and accuracy-efficiency frontier of RIFT.} Left: RIFT prioritizes morphology-relevant cues over generic feature mixing, including local structure, cooperative direction, and controlled receptive fields. Right: average mIoU versus CUDA FPS, with circle radius denoting parameter count. RIFT-T and RIFT-B occupy the upper-right frontier, indicating strong accuracy-efficiency trade-offs.}
    \label{fig:1teaser}
\end{figure}

Crack segmentation~\cite{benz2024omnicrack30k, lin2023deepcrackat} is fundamental to intelligent infrastructure inspection, with applications in roads, bridges, tunnels, and concrete health monitoring. Compared with classification or detection, pixel-level segmentation provides finer structural cues, including crack location, boundary, width, and connectivity, which are essential for measurement, severity assessment, and maintenance decisions. However, cracks are often low-contrast, thin, sparse, locally discontinuous, and morphologically irregular. They occupy small foreground regions and are easily confused with textures, joints, shadows, and stains~\cite{joo2026frequency}. Crack segmentation should therefore be viewed not as conventional semantic region discrimination, but as sparse structural recovery under severe visual ambiguity. 

Recent crack segmentation methods largely follow general semantic segmentation~\cite{guo2018review}, introducing stronger convolutional backbones~\cite{woo2023convnext}, multi-scale fusion, boundary refinement, Transformers~\cite{vaswani2017attention}, Mamba-style state-space models~\cite{liu2025vision}, and hybrid encoders to improve representation capacity and contextual modeling. While improving accuracy, these designs also increase architectural complexity, memory consumption, and deployment cost (Figure~\ref{fig:1teaser}). More importantly, they do not directly clarify which cues are critical for crack recovery. Unlike common semantic objects, cracks have limited category-level semantics but strong sparse structural regularity. Their discriminative evidence mainly lies in fine edge traces, anisotropic propagation, local continuity, branching patterns, and topology preservation, rather than high-level object semantics or global scene understanding.

This mismatch motivates a direct question: \textit{should efficient crack segmentation rely on increasingly complex generic feature mixing, or can a simpler task-aligned formulation be equally, or even more, effective?} We argue for the latter. Since cracks are sparse, elongated, locally discontinuous, and directionally organized rather than conventional semantic regions, limited capacity should be allocated to the cues that govern crack recovery: weak local structural evidence, directional continuity, and controlled receptive fields. These cues are not auxiliary refinements to generic representation learning, but direct responses to the major failure modes of sparse crack recovery.

Based on this principle, we propose \textbf{RIFT}, a compact crack segmentation family centered on structural and directional modeling. RIFT adopts a lightweight encoder-decoder architecture with two coupled components. The Structural and Directional Modeling block preserves local texture, edge, and geometric traces while aggregating horizontal, vertical, and oblique continuity responses for elongated and fragmented cracks. Its cooperative directional aggregation represents bending, branching, and interweaving structures without enforcing a single dominant direction. A lightweight multi-scale structural decoder then restores high-resolution masks by using high-level structural context to guide low-level detail injection. Thus, RIFT is not a compressed generic segmentation model, but a deliberately compact formulation that allocates capacity to morphology-relevant crack cues.

Experiments on DeepCrack~\cite{liu2019deepcrack}, CamCrack789~\cite{zhu2024lightweight}, CrackMap~\cite{katsamenis2023few}, and Crack500~\cite{yang2019feature} validate this task-aligned design. RIFT achieves the best or tied-best results across the 16 main metrics, with RIFT-B giving the strongest accuracy and RIFT-T the best deployment efficiency. Topology-aware evaluation further shows improved skeleton-region consistency, centerline reconstruction, and connectivity preservation. Ablations attribute the gains mainly to local structural and directional continuity modeling, while extra frequency or context enhancement brings no consistent improvement. Cross-dataset transfer and visualizations suggest that RIFT captures transferable crack structures rather than dataset-specific textures. Overall, morphology-aligned simplicity yields a stronger accuracy-efficiency trade-off than heavier generic feature-mixing designs.

The main contributions are summarized as follows:
\begin{itemize}
    \item We reformulate efficient crack segmentation as sparse structural recovery, emphasizing weak local evidence, directional continuity, and background decoupling over high-level semantic enrichment.
    \item We propose RIFT, a compact morphology-aligned model family built on local structural preservation, cooperative directional aggregation, and lightweight structural recovery.
    \item Experiments on four benchmarks demonstrate strong accuracy, topology preservation, transfer robustness, and deployment efficiency, showing that morphology-aligned simplicity can outperform more complex designs for sparse crack structures.
\end{itemize}

\section{Related Work}\label{sec2}

\textbf{CNN-based crack segmentation.}
Early crack segmentation methods largely adopted encoder-decoder designs, with U-Net~\cite{ronneberger2015u} and UNet++~\cite{zhou2019unet++} as common backbones. For thin, sparse, and low-contrast cracks, hierarchical edge cues and multi-scale textures are crucial, as shown by HED~\cite{xie2015holistically}, DeepCrack~\cite{liu2019deepcrack}, and Chen et al.~\cite{chen2023automatic}. CNN-based methods remain strong baselines~\cite{zhou2023deep} because convolutional locality matches fine crack structures. However, most still rely on generic local aggregation and lack explicit modeling of anisotropic continuity.

\noindent\textbf{Hybrid global modeling and foundation adaptation.}
Recent methods strengthen global modeling for long-range propagation and discontinuous connectivity, including CNN-Transformer hybrids such as DTrCNet~\cite{xiang2023crack}, Wang et al.~\cite{wang2024dual}, and PCTC-Net~\cite{moon2024pctc}. Foundation adaptation has also been explored through fine-tuned vision foundation models~\cite{ge2024fine} and SAM-style paradigms~\cite{rostami2026segment}. Although these methods improve contextual reasoning, they increase architectural complexity and heterogeneous feature fusion. For cracks dominated by sparse local traces and anisotropic continuity, generic global mixing may not allocate capacity most efficiently.

\noindent\textbf{Efficient and state-space crack segmentation.}
Efficiency is critical for infrastructure inspection under limited computation and memory. CarNet~\cite{li2024rethinking} and MixerCSeg~\cite{zhao2026mixercseg} study accuracy-efficiency trade-offs, while Vision Mamba has been introduced into crack segmentation~\cite{chen2024vision}, with SCSegamba~\cite{liu2025scsegamba} combining structure-aware design and lightweight scanning. OmniCrack30k~\cite{benz2024omnicrack30k} further promotes large-scale evaluation across materials and scenarios. Unlike methods that mainly lighten generic modules or compress feature mixing, RIFT starts from crack morphology and examines the structure-specific modeling needed for efficient recovery.

\noindent\textbf{Structure- and topology-aware modeling.}
Fine structures and topological continuity have also been studied. HED~\cite{xie2015holistically} and DeepCrack~\cite{liu2019deepcrack} highlight edge-aware recovery. RINDNet~\cite{pu2021rindnet} suggests that structural targets cannot rely on regional semantics alone, while clDice~\cite{shit2021cldice} shows the limits of pixel-overlap metrics for topology-sensitive structures. SCSegamba~\cite{liu2025scsegamba} also emphasizes morphology- and topology-aware perception. RIFT follows this direction but distills it into a simple local-directional inductive bias for sparse crack recovery.

Overall, existing studies improve crack segmentation through stronger backbones, global mixing, efficient modules, and topology-aware objectives. We instead ask whether a smaller morphology-aligned cue set is sufficient. By prioritizing local structure, directional continuity, and receptive-field control, RIFT can match or surpass more complex designs with lower deployment cost.

\begin{figure*}
    \centering
    \includegraphics[width=0.85\linewidth]{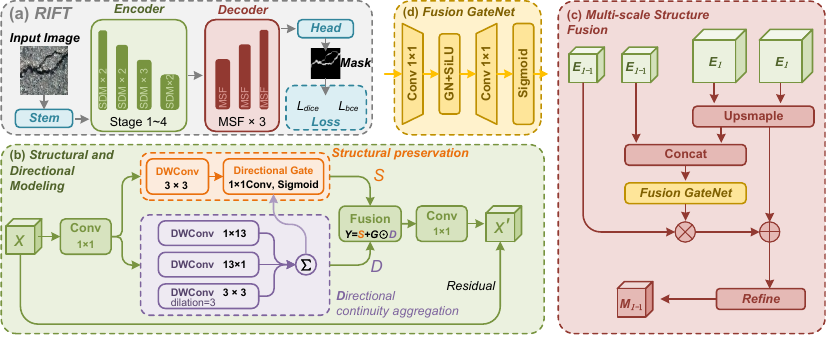}
    \caption{\textbf{Overall architecture of RIFT.} (a) RIFT consists of a stem, four-stage encoder, multi-scale decoder, and prediction head. Block numbers indicate RIFT-B, with the reduced-depth RIFT-T detailed in \textbf{Appendix~A.1}. (b) The Structural and Directional Modeling block preserves local structure and gates directional continuity. (c) The Multi-scale Structure Fusion module recovers high-resolution cracks through gated lateral fusion. (d) GateNet predicts adaptive fusion weights from concatenated decoder features.}
    \label{fig:2rift}
\end{figure*}

\section{Task Formulation and Design Rationale}\label{sec3}
Given an image $I\in \mathbb{R}^{3\times H\times W}$, crack segmentation predicts a binary mask $M\in\{0,1\}^{H\times W}$. While it can be posed as pixel-level binary classification~\cite{Hussain2025PixelLevel}, this formulation hides its structural nature. Cracks occupy small foreground regions and are low-contrast, thin, discontinuous, bending, and branching~\cite{Pereira2025SkeletonLoss}; thus, the task is better viewed as recovering sparse, elongated, and locally continuous structures from ambiguous evidence.

\textbf{Failure modes.}
Crack segmentation faces three structural failure modes. Local ambiguity arises when cracks resemble joints, texture edges, scratches, shadows, or stains~\cite{Wang2025EfficientSegNet}. Structural breakage occurs when low contrast, occlusion, or noise interrupts crack continuity. Background coupling appears when enlarged receptive fields or excessive contextual mixing connect crack-like background structures with true cracks~\cite{Zhang2024SCSNet}. Thus, stronger generic representation alone is insufficient; the model must preserve weak crack evidence without amplifying non-crack structures.

\textbf{Design principles.}
These failure modes motivate fine local structural evidence for local ambiguity, directional continuity modeling for structural breakage, and controlled anisotropic receptive fields for background coupling~\cite{Qu2025DCFEN}. RIFT therefore prioritizes local structure and directional continuity while using controlled receptive fields to expand structural support without excessive background interference.

\textbf{Cooperative direction aggregation.}
Real cracks rarely follow a single dominant direction. In bending, branching, or interweaving regions, multiple directions may be informative simultaneously. Let $p$ be a spatial position, $i\in\{1,\dots,d\}$ a direction index, and $e_i(p)$ the response along direction $i$. A competitive aggregation can be written as:
\begin{equation}
    E_{\mathrm{comp}}(p)=\sum_{i=1}^{d}\alpha_i(p)e_i(p),
    \quad \sum_{i=1}^{d}\alpha_i(p)=1,
\end{equation}
where $\alpha_i(p) \ge 0$ is the normalized weight. As a convex combination, it may suppress non-dominant but useful directional cues. In contrast, additive aggregation is defined as:
\begin{equation}
E_{\mathrm{add}}(p)=\sum_{i=1}^{d}e_i(p).
\end{equation}
It does not force one direction to dominate, making it suitable for bending, branching, and interweaving cracks where several directional responses jointly support prediction. Thus, RIFT preserves cooperative directional evidence rather than selecting only the most salient direction.

\textbf{Controlled directional receptive field.}
Directional continuity requires sufficiently large anisotropic receptive fields, but overly large kernels may absorb textures, joints, shadows, and other crack-like distractors. Let $k$ denote the directional kernel size. Conceptually, its effect can be written as a trade-off between continuity benefit $\mathcal{C}(k)$ and background coupling cost $\mathcal{N}(k)$:
\begin{equation}
\mathcal{R}(k)=\mathcal{C}(k)-\mathcal{N}(k),
\end{equation}
where $\mathcal{C}(k)$ tends to saturate with increasing $k$, while $\mathcal{N}(k)$ may keep growing once distractors enter the receptive field. Crack segmentation therefore favors moderate anisotropic kernels that balance continuity modeling and interference suppression. The following method instantiates these principles with a compact structural-directional block and a lightweight multi-scale structural decoder, and empirically studies $k$ in the ablation.

\section{Method}\label{sec4}
RIFT instantiates the above principles in a compact encoder-decoder architecture. As shown in Figure~\ref{fig:2rift}, the encoder stacks Structural and Directional Modeling Blocks (SDMs) to preserve local evidence and aggregate crack-continuity cues, while the decoder restores high-resolution crack structures through lightweight multi-scale fusion.

Given an input image $I \in \mathbb{R}^{3 \times H \times W}$, the stem extracts low-level features and reduces spatial resolution. A four-stage encoder then produces hierarchical features with SDMs. These features are fused by Multi-scale Structure Fusion (MSF) modules, and the prediction head outputs pixel-level logits $P \in \mathbb{R}^{1 \times H \times W}$. Figure~\ref{fig:2rift}(a) shows the framework, while Figures~\ref{fig:2rift}(b)--(d) illustrate the SDM, MSF module, and GateNet. Overall, RIFT is a task-shaped compact formulation that allocates capacity to local structure, directional continuity, and structural recovery.

\subsection{Structural and Directional Modeling}\label{sec4.2}
SDM is the core unit that translates the design rationale into lightweight operators. It preserves local structural evidence to reduce ambiguity, aggregates directional continuity cues to recover interrupted propagation, and modulates directional responses to suppress background coupling.

Given an input feature $X \in \mathbb{R}^{C \times H \times W}$, SDM first applies normalization and channel expansion:
\begin{equation}
\hat{X} = \mathrm{Conv}_{1\times1}(\mathrm{Norm}(X)),
\end{equation}
where $\mathrm{Norm}(\cdot)$ denotes normalization and $\mathrm{Conv}_{1\times1}$ projects the input into a hidden space for joint structural-directional modeling.

\textbf{Local structural preservation.}
To address local ambiguity, SDM first extracts local structural evidence before introducing larger directional aggregation. A depth-wise $3\times3$ convolution $S = f_\mathrm{dw}^{3\times3}(\hat{X})$ is applied to $\hat{X}$, $S$ denotes the local structural response. This branch preserves fine-grained texture, and edge traces at low computational cost, providing the local evidence required to distinguish weak cracks from joints, shadows, and other crack-like distractors.

\textbf{Directional continuity aggregation.}
Local evidence alone is insufficient when cracks are low-contrast, elongated, or locally broken. SDM therefore aggregates responses along several complementary directional patterns:
\begin{equation}
D_h = f_\mathrm{dw,h}^{1\times k}(\hat{X}), 
D_v = f_\mathrm{dw,v}^{k\times 1}(\hat{X}), 
D_d = f_\mathrm{dw,d}^{3\times 3}(\hat{X}),
\end{equation}
where $f_\mathrm{dw,h}^{1\times k}$ and $f_\mathrm{dw,v}^{k\times1}$ denote horizontal and vertical depth-wise anisotropic convolutions, respectively, and $f_\mathrm{dw,d}^{3\times3}$ denotes a dilated depth-wise convolution that provides a lightweight approximation of oblique local continuity. Following the cooperative aggregation principle, the directional response is computed as:
\begin{equation}
D = D_h + D_v + D_d.
\end{equation}
This additive form keeps multiple directional cues active and is therefore suitable for cracks with bending, branching, or interweaving patterns. The kernel size $k$ controls the anisotropic receptive field, and its effect is analyzed in the ablation study.

\textbf{Structure-aware directional modulation.}
Directional aggregation may also respond to crack-like background textures. SDM therefore constrains directional information using the joint structural-directional state. A lightweight gate is generated from the fused response:
\begin{equation}
G = \sigma \left(f_g(S + D)\right),
\end{equation}
where $f_g(\cdot)$ denotes a lightweight gating mapping and $\sigma(\cdot)$ denotes the Sigmoid function. The gated fusion is formulated as:
\begin{equation}
Y = S + G \odot D,
\end{equation}
where $\odot$ denotes element-wise multiplication. In this formulation, directional responses contribute more when supported by local structural evidence, thereby reducing the risk of propagating background distractors. Finally, the fused feature is projected back to the original channel dimension and added to the input:
\begin{equation}
\tilde{Y} = f_p(Y), \qquad
X' = X + \tilde{Y},
\end{equation}
where $f_p(\cdot)$ denotes the output projection. SDM thus provides a compact collaboration among local evidence, directional continuity, and background suppression.

\begin{table*}
\centering
\setlength\tabcolsep{1.2mm}
\caption{\textbf{Quantitative comparison.} All metrics are percentages (\%). Except for $\dagger$ methods adopted from~\cite{zhang2026MTSCrack} for reference, all competitors are reproduced under the MixerCSeg evaluation setting and reported from a single run. RIFT-T and RIFT-B are evaluated over three random seeds, with mean results reported and \textit{std} shown in the last two rows. Best and second-best results are in \textbf{bold} and \underline{underline}.}
\label{tab:1main results}
\begin{tabular}{lcccc|cccc|cccc|cccc}
\toprule
\multirow{2}{*}{\textbf{Method}} & \multicolumn{4}{c|}{\textbf{DeepCrack}} & \multicolumn{4}{c|}{\textbf{CamCrack789}} & \multicolumn{4}{c|}{\textbf{CrackMap}} & \multicolumn{4}{c}{\textbf{Crack500}} \\ \cmidrule(l){2-17} 
 & \textbf{mIoU} & \textbf{ODS} & \textbf{OIS} & \textbf{F1} & \textbf{mIoU} & \textbf{ODS} & \textbf{OIS} & \textbf{F1} & \textbf{mIoU} & \textbf{ODS} & \textbf{OIS} & \textbf{F1} & \textbf{mIoU} & \textbf{ODS} & \textbf{OIS} & \textbf{F1} \\ \midrule
MTSCrack$^\dagger$ & 83.6 & - & - & 81.7 & - & - & - & - & - & - & - & - & 75.8 & - & - & 73.4 \\
RINDNet & 83.9 & 80.9 & 82.7 & 83.8 & 81.4 & 76.3 & 77.3 & 60.5 & 74.3 & 67.5 & 69.4 & 67.0 & 73.8 & 64.7 & 64.8 & 71.2 \\
DTrCNet & 86.6 & 84.7 & 85.1 & 85.7 & 81.5 & 77.2 & 77.9 & 78.4 & 78.1 & 73.3 & 74.1 & 72.8 & 76.2 & 70.1 & 72.4 & 73.6 \\
CarNet & 88.8 & 86.9 & 87.0 & 89.4 & 82.6 & 78.3 & 78.5 & 72.5 & 75.7 & 68.4 & 71.2 & 64.0 & 64.3 & 54.9 & 59.2 & 36.4 \\
RestorMixer & 90.1 & 88.9 & 89.6 & 90.5 & 83.6 & 79.7 & 80.2 & 73.9 & 78.9 & 74.1 & 78.4 & 66.3 & 74.2 & 66.4 & 68.2 & 69.6 \\
SCSegamba & 90.2 & 89.4 & 89.9 & 91.1 & 82.7 & 79.2 & 79.4 & 78.8 & 80.9 & 77.4 & 77.7 & 76.8 & 77.7 & 72.4 & 73.7 & 75.5 \\
MixerCSeg & 91.5 & 90.9 & 92.0 & 92.1 & 84.1 & 81.2 & 82.0 & 82.4 & 81.2 & 77.8 & 78.1 & \underline{78.2} & 78.2 & 72.8 & 74.8 & 77.6 \\
RIFT-T (ours) & \underline{92.0} & \underline{91.5} & \textbf{92.2} & \underline{92.4} & \underline{85.6} & \underline{83.2} & \underline{84.1} & \underline{84.4} & \underline{81.2} & \underline{77.8} & \underline{78.7} & 77.5 & \underline{79.2} & \underline{74.3} & \underline{75.5} & \underline{78.4} \\
RIFT-B (ours) & \textbf{92.0} & \textbf{91.5} & \underline{92.0} & \textbf{92.7} & \textbf{86.1} & \textbf{83.9} & \textbf{84.9} & \textbf{85.2} & \textbf{82.7} & \textbf{80.0} & \textbf{80.6} & \textbf{80.0} & \textbf{79.4} & \textbf{74.7} & \textbf{76.2} & \textbf{78.7} \\ \midrule
RIFT-T \textit{std} & 0.16 & 0.17 & 0.12 & 0.44 & 0.07 & 0.12 & 0.15 & 0.20 & 0.29 & 0.40 & 0.29 & 0.75 & 0.09 & 0.18 & 0.50 & 0.05 \\
RIFT-B \textit{std} & 0.27 & 0.36 & 0.32 & 0.32 & 0.08 & 0.10 & 0.16 & 0.15 & 0.53 & 0.72 & 0.83 & 0.63 & 0.09 & 0.14 & 0.41 & 0.11 \\ \bottomrule
\end{tabular}
\end{table*}

\subsection{Multi-scale Structural Decoder}\label{sec4.3}
After hierarchical encoding, RIFT obtains multi-scale structure-direction representations. The decoder is designed as a lightweight structural recovery module, not a heavy semantic enhancement head. It recovers high-resolution crack structures while selectively injecting low-level details through a top-down design built from MSF.

Let the encoder feature at level $l$ be denoted as $E_l$, where $l\in\{1,2,3,4\}$. Since different levels have inconsistent channel dimensions, they are first projected into a unified decoder space by $F_l = g_l(E_l)$, where $g_l(\cdot)$ denotes a $1\times1$ projection. Let $H_l$ denote the decoded feature at level $l$, with $H_4 = F_4$ as the initialization. For each top-down step from level $l$ to level $l-1$, the higher-level feature is upsampled as $\hat{H}_{l-1} = \mathrm{Up}(H_l)$, where $\mathrm{Up}(\cdot)$ denotes bilinear upsampling. This operation propagates higher-level structural context and directional continuity to the current resolution.

MSF then uses GateNet to generate an adaptive fusion weight from the upsampled feature and the current lateral feature:
\begin{equation}
A_{l-1} = \mathrm{GateNet}\left([\hat{H}_{l-1}, F_{l-1}]\right),
\end{equation}
where $[\cdot,\cdot]$ denotes channel concatenation. The gated fusion is defined as:
\begin{equation}
\tilde{H}_{l-1} = \hat{H}_{l-1} + A_{l-1} \odot F_{l-1}.
\end{equation}
This operation allows low-level edge and texture details to be introduced under higher-level structural guidance, thereby reducing interference from noisy or pseudo-structural responses. A lightweight refinement module is then applied:
\begin{equation}
H_{l-1} = r_{l-1}(\tilde{H}_{l-1}),
\end{equation}
where $r_{l-1}(\cdot)$ denotes convolutional refinement. After the final decoding stage, the prediction head outputs logits as $P = f_{\mathrm{head}}(H_1)$, where $P \in \mathbb{R}^{1 \times H \times W}$. Overall, the decoder complements SDM by restoring spatial details without introducing a heavy semantic fusion head.

\subsection{Variants and Efficiency}\label{sec4.4}
RIFT keeps the same architecture across model scales, varying only the encoder width, encoder depth, and decoder dimension. We instantiate RIFT-T for efficient deployment and RIFT-B as the default base model. For a fair comparison, RIFT is trained with the same BCE-Dice objective as MixerCSeg~\cite{zhao2026mixercseg}:
\begin{equation}
    \mathcal{L} = \lambda_1 \mathcal{L}_{\mathrm{bce}} + \lambda_2 \mathcal{L}_{\mathrm{dice}},
\end{equation}
where $\lambda_1 : \lambda_2 = 5 : 1$. Since RIFT is built from lightweight convolutions, additive directional aggregation, and gated multi-scale fusion, it controls not only the parameter count but also activation memory and inference latency. More implementation details are provided in \textbf{Appendix A}.

\section{Experiments}\label{sec5}

\subsection{Experimental Settings}
We evaluate RIFT on DeepCrack, CamCrack789, CrackMap, and Crack500, and compare it with representative CNN-based, hybrid, state-space, and lightweight methods: LMM~\cite{al2024LMM}, EfficientCrackNet~\cite{zim2025efficientcracknet}, MTSCrack~\cite{zhang2026MTSCrack}, CarNet~\cite{li2024rethinking}, RINDNet~\cite{pu2021rindnet}, DTrCNet~\cite{xiang2023crack}, SCSegamba~\cite{liu2025scsegamba}, RestorMixer~\cite{gu2025efficient}, and MixerCSeg~\cite{zhao2026mixercseg}. Except for $\dagger$ methods adopted from~\cite{zhang2026MTSCrack} as references, all competitors are reproduced under the MixerCSeg setting. RIFT-T and RIFT-B use $512 \times 512$ inputs, BCE-Dice loss with $\lambda_1:\lambda_2=5:1$, and the same evaluation protocol. The directional kernel size is $13$ unless otherwise specified. Training details and hardware are in \textbf{Appendix B.1--B.3}.

We report mIoU, ODS, OIS, and fixed-threshold F1 as main region-level metrics. RIFT variants are evaluated over three random seeds and reported by mean and \textit{std}, while reproduced competitors use a single run unless otherwise specified. For structural recovery, we also report clDice~\cite{shit2021cldice}, Skeleton F1, and Connectivity Ratio (CR), with definitions in \textbf{Appendix B.4}.

\subsection{Main Results}

\textbf{Quantitative comparison.}
Table~\ref{tab:1main results} shows that RIFT performs strongly across all four benchmarks, achieving the best or tied-best results among the 16 main metrics against reproduced baselines and external references. RIFT-B provides the strongest overall accuracy, leading on CamCrack789, CrackMap, and Crack500, while RIFT-T and RIFT-B jointly perform best on DeepCrack. RIFT-T remains competitive with substantially better deployment efficiency. These results indicate that a compact structural-directional design can match or surpass complex hybrid feature-mixing architectures when aligned with crack morphology, yielding a favorable accuracy-efficiency trade-off for sparse structural segmentation. The small standard deviations over three seeds show stable optimization, with training curves provided in \textbf{Appendix D.4}.

\begin{table}[t]
\centering
\caption{\textbf{Topology-aware evaluation (\%).} clDice, Skeleton F1, and CR evaluate skeleton-region consistency, centerline reconstruction, and connectivity preservation. clDice and Skeleton F1 are averaged over four datasets, while CR uses the three datasets with available connectivity statistics.}
\label{tab:2topology aware}
\begin{tabular}{lccc}
\toprule
\textbf{Method} & \textbf{clDice$\uparrow$} & \textbf{Skeleton F1$\uparrow$} & \textbf{CR$\uparrow$} \\ \midrule
SCSegamba & 86.0 & 83.0 & 10.5 \\
MixerCSeg & 89.1 & 84.8 & 9.5 \\
RIFT-T (ours) & \underline{90.1} & \underline{85.7} & \textbf{12.3} \\
RIFT-B (ours) & \textbf{90.6} & \textbf{86.5} & \underline{11.8} \\ \bottomrule
\end{tabular}
\end{table}

\begin{table}
\centering
\setlength\tabcolsep{1mm}
\caption{\textbf{Efficiency comparison.} Comparison of FLOPs (G)$\downarrow$, parameters (M)$\downarrow$, peak memory usage (MiB)$\downarrow$, and FPS$\uparrow$ among different methods. $\dagger$ denotes externally reported results at $256 \times 256$, included only for reference.}
\label{tab:3efficiency}
\begin{tabular}{lcccc}
\toprule
\textbf{Method} & \textbf{FLOPs} & \textbf{Params} & \textbf{Memory} & \textbf{FPS} \\ \midrule
CarNet & 19.00 & 4.89 & 1824 & 82 \\
SCSegamba & 18.16 & 2.80 & 2206 & 13 \\
MixerCSeg & \underline{2.05} & 2.54 & 1190 & 4 \\
LMM$^\dagger$ & 6.56 & 0.87 & - & 18 \\
EfficientCrackNet$^\dagger$ & 3.00 & \underline{0.28} & - & 13 \\
MTSCrack$^\dagger$ & 2.63 & \textbf{0.19} & - & \underline{180} \\
RIFT-T (ours) & \textbf{1.39} & 0.47 & \textbf{268} & \textbf{355} \\
RIFT-B (ours) & 3.34 & 1.69 & \underline{341} & 149 \\ \bottomrule
\end{tabular}
\end{table}

\textbf{Structural recovery evaluation.}
Beyond region-level overlap, Table~\ref{tab:2topology aware} evaluates whether RIFT preserves crack structures more effectively. RIFT-B achieves the best clDice and Skeleton F1, indicating stronger skeleton-region consistency and centerline reconstruction. For CR, which sets a stricter criterion for preserving connected crack segments, RIFT-T obtains the best result and RIFT-B ranks second. These results are important because they support the main thesis of this paper: the improvement is not merely better mask filling, but more accurate recovery of the sparse structural properties that define thin cracks. Per-dataset image-level and dataset-level topology-aware statistics are provided in \textbf{Appendix D.1}.

\textbf{Efficiency comparison.}
Table~\ref{tab:3efficiency} compares FLOPs, parameters, peak memory, and FPS. RIFT-T is the most efficient, requiring only 1.39G FLOPs, 0.47M parameters, 268 MiB peak memory, and reaching 355 FPS. RIFT-B remains compact with 3.34G FLOPs, 1.69M parameters, 341 MiB peak memory, and 149 FPS. Compared with SCSegamba and MixerCSeg, RIFT uses less memory and achieves much higher throughput, confirming that morphology-aligned simplicity supports both segmentation quality and deployment efficiency. FLOPs alone do not explain practical behavior, as RIFT-B is faster and more memory-efficient than MixerCSeg despite slightly higher FLOPs. $\dagger$ results use a different input resolution and are included only as references. See \textbf{Appendix B.3}.

\subsection{Ablation and Further Analysis}

\textbf{Component analysis.}
Table~\ref{tab:4ablation} shows that directional modeling improves the Local-only baseline mainly in F1 and yields larger gains with the gated decoder, especially on CrackMap. The gated decoder alone does not consistently improve performance, indicating that adaptive detail injection is most useful when direction-aware structural features are already provided by the encoder. Frequency enhancement and global context modeling bring no consistent gains; the degradation of RIFT-B + Freq. suggests that direct high-frequency amplification may also strengthen background textures and crack-like noise. These results show that the improvement does not come from adding enhancement branches, but from a compact structural-directional design that directly targets the dominant failure modes of crack segmentation.

\begin{table}
\centering
\setlength\tabcolsep{1.2mm}
\caption{\textbf{Component ablation in RIFT-B.} Local-only uses only the local structural branch and a plain decoder. Gated Dec., Directional, Freq., and Context denote the gated decoder, directional branch, frequency-enhancement branch, and global context branch, respectively.}
\label{tab:4ablation}
\begin{tabular}{lcccc}
\toprule
\multirow{2}{*}{\textbf{Method}} & \multicolumn{2}{c}{\textbf{CrackMap}} & \multicolumn{2}{c}{\textbf{DeepCrack}} \\ \cmidrule(l){2-5} 
 & \textbf{mIoU} & \textbf{F1} & \textbf{mIoU} & \textbf{F1} \\ \midrule
Local-only & 82.5 & 79.4 & 92.1 & 92.1 \\
Local + Gated Dec. & 82.1 & 79.3 & 91.9 & 92.8 \\
Directional + Plain Dec. & 82.5 & 79.9 & 92.1 & \textbf{92.9} \\
RIFT-B & \textbf{83.0} & \textbf{80.3} & \textbf{92.1} & 92.7 \\ \midrule
RIFT-B + Freq. & 81.4 & 76.2 & 91.4 & 92.1 \\
RIFT-B + Context & 82.6 & 80.0 & 91.6 & 92.3 \\
RIFT-B + Freq. + Context & 82.2 & 79.5 & 91.9 & 92.2 \\ \bottomrule
\end{tabular}
\end{table}

\begin{table}
\centering
\caption{\textbf{Directional aggregation variants in RIFT-B.} SoftDir-Comp. replaces additive directional aggregation with competitive soft directional selection. Multi-Diag. adds an additional dilated oblique-continuity branch.}
\label{tab:5directional aggregation}
\begin{tabular}{lcccc}
\toprule
\textbf{Method} & \textbf{mIoU} & \textbf{ODS} & \textbf{OIS} & \textbf{F1} \\ \midrule
SoftDir-Comp. & 80.8 & 77.3 & 77.6 & 77.3 \\
Multi-Diag. & 82.0 & 78.9 & 79.4 & 79.0 \\
RIFT-B & \textbf{83.0} & \textbf{80.5} & \textbf{81.3} & \textbf{80.3} \\ \bottomrule
\end{tabular}
\end{table}

\textbf{Direction aggregation strategy.}
Table~\ref{tab:5directional aggregation} validates the cooperative direction aggregation principle. Competitive soft selection consistently underperforms additive aggregation across all four metrics, showing that suppressing non-dominant directions harms complex crack recovery. Adding another diagonal branch also degrades performance, indicating that more directional operators are not necessarily beneficial. Thus, the default additive design is not merely an implementation choice: cooperative multi-directional support compactly represents bending, branching, and interwoven cracks, while extra branches may be redundant or counterproductive.

\begin{figure}
    \centering
    \includegraphics[width=0.75\linewidth]{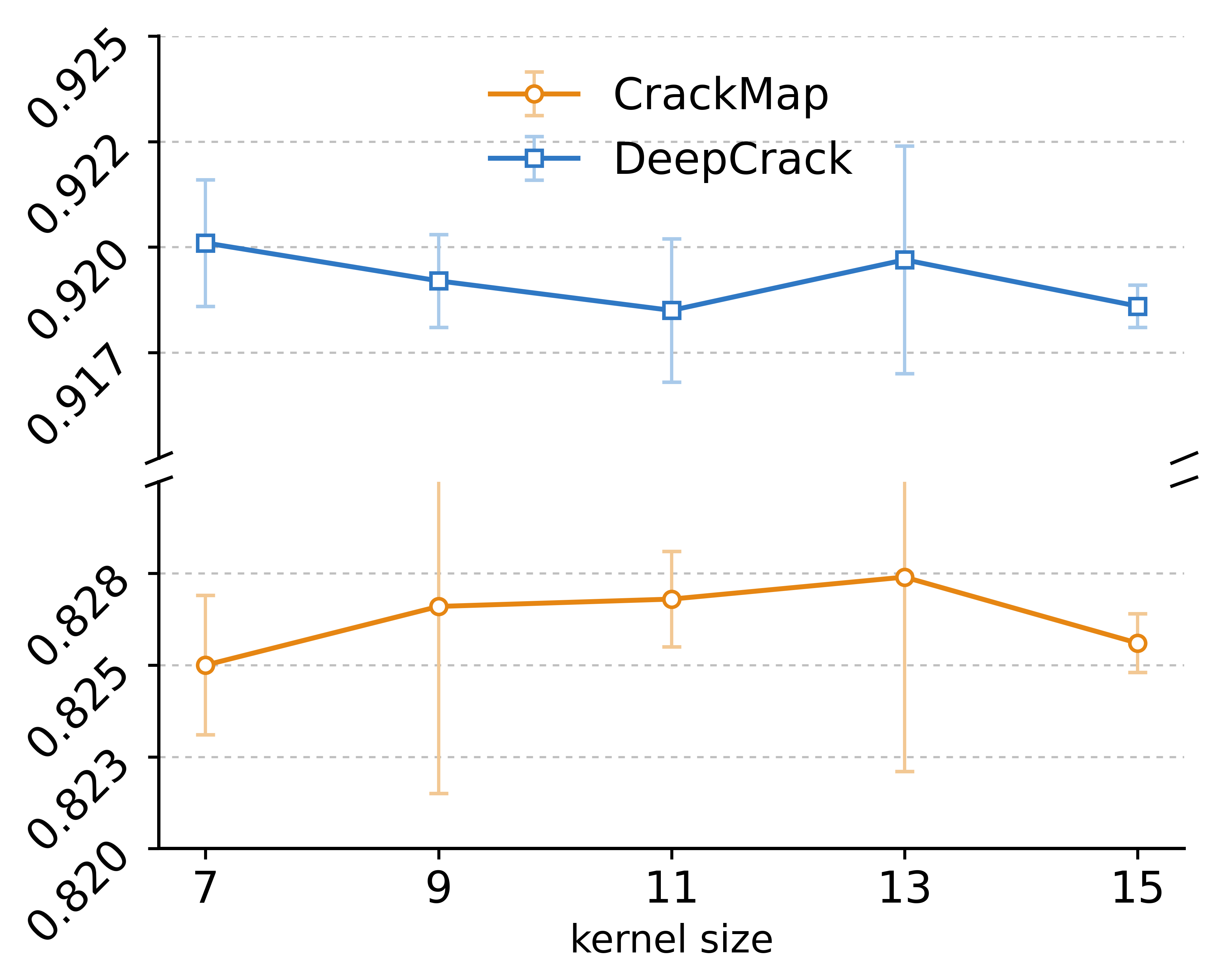}
    \caption{\textbf{Ablation of kernel size in RIFT-B.} Markers and error bars denote the \textit{mean} mIoU and \textit{std} over random seeds.}
    \label{fig:3kernel size}
\end{figure}

\textbf{Directional kernel analysis.}
Figure~\ref{fig:3kernel size} shows that the performance on CrackMap improves up to $k=13$ and decreases at $k=15$, while medium-sized kernels remain consistently competitive on DeepCrack. This observation supports the controlled receptive-field principle: larger kernels help recover directional continuity within an appropriate range, whereas overly large kernels may introduce irrelevant structural coupling, as reflected by the performance drop at $k=15$. Therefore, $k=13$ is used in all main experiments. Additional stage-wise kernel analysis is provided in \textbf{Appendix C.1}.

\begin{table}
\centering
\setlength\tabcolsep{1.5mm}
\caption{\textbf{Representative cross-dataset transfer results (mIoU, \%).} Column groups are training sources, and sub-columns are test targets without fine-tuning. Avg. is the row average. C789 and CMap denote CamCrack789 and CrackMap. See \textbf{Appendix D.3} for more results.}
\label{tab:6cross dataset}
\begin{tabular}{lccccc}
\toprule
\multirow{2}{*}{\textbf{Method}} & \multicolumn{2}{c}{\textbf{DeepCrack}} & \multicolumn{2}{c}{\textbf{Crack500}} & \multirow{2}{*}{\textbf{Avg.}} \\ \cmidrule(l){2-5}
 & \textbf{C789} & \textbf{CMap} & \textbf{C789} & \textbf{CMap} &  \\ \midrule
CarNet & 75.6 & 57.5 & 65.5 & 72.2 & 67.7 \\
RINDNet & 73.6 & 60.2 & 68.7 & 75.3 & 69.5 \\
DTrCNet & 74.9 & 63.6 & 70.5 & 78.0 & 71.8 \\
SCSegamba & 78.9 & 68.5 & 73.8 & 79.6 & 75.2 \\
MixerCSeg & 78.2 & 60.5 & 72.2 & 81.3 & 73.1 \\
RIFT-T (ours) & \textbf{81.7} & \textbf{73.8} & \textbf{77.9} & \textbf{82.9} & \textbf{79.1} \\
RIFT-B (ours) & \underline{81.3} & \underline{68.8} & \underline{76.7} & \underline{81.5} & \underline{77.1} \\ \bottomrule
\end{tabular}
\end{table}

\textbf{Cross-dataset generalization.}
Table~\ref{tab:6cross dataset} reports direct transfer without target-domain fine-tuning, testing whether models capture transferable crack morphology rather than dataset-specific textures or annotations. In the representative settings, RIFT-T and RIFT-B achieve the strongest average performance, reaching 79.1 and 77.1 mIoU, respectively. The more compact RIFT-T transfers better than RIFT-B, suggesting that the gain is not merely capacity-driven and that a tighter structural-directional bias can improve robustness to domain appearance. This supports the practical value of task-aligned simplicity: a smaller model can reduce deployment cost while generalizing better when task-relevant cues are compactly modeled. The full matrix in \textbf{Appendix~D.3} confirms RIFT's best overall transfer averages across all source-target pairs.

\begin{figure}
    \centering
    \includegraphics[width=0.85\linewidth]{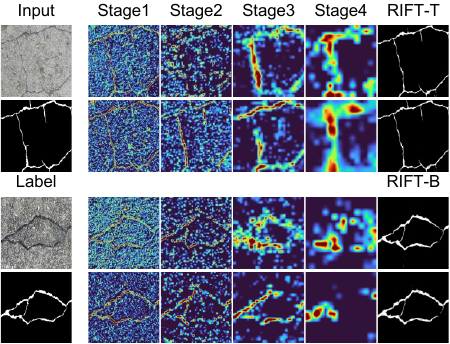}
    \caption{\textbf{Stage-wise feature visualization.} Normalized responses show how encoder features evolve from edge-texture cues to crack-structure activations.}
    \label{fig:4stage-wise feature}
\end{figure}

\textbf{Stage-wise feature visualization.}
Figure~\ref{fig:4stage-wise feature} visualizes normalized stage-wise feature responses rather than final predictions. Both RIFT-T and RIFT-B show a clear hierarchy: early stages mainly respond to local textures and edges, while deeper stages focus more on crack trunks, branches, and connectivity-related regions. Compared with RIFT-T, RIFT-B produces more concentrated and structurally complete middle- and high-stage responses, consistent with its stronger quantitative performance.

\begin{figure}
    \centering
    \includegraphics[width=1\linewidth]{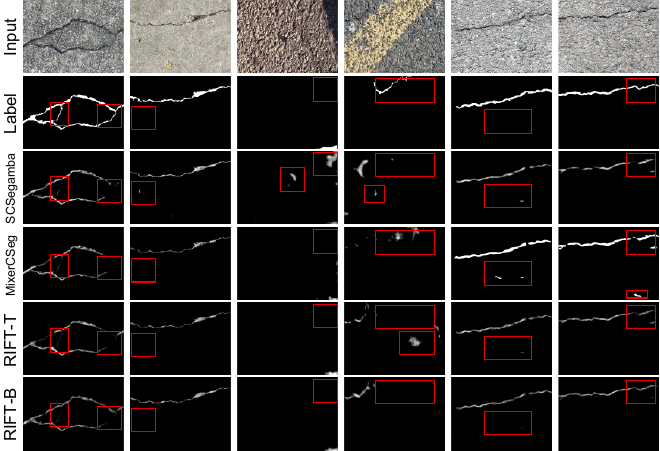}
    \caption{\textbf{Qualitative comparison.} Red boxes mark differences in thin-crack continuity, branching recovery, and background suppression. More results are in \textbf{Appendix D.5}.}
    \label{fig:5vis comparison}
\end{figure}

\textbf{Qualitative comparison.}
Figure~\ref{fig:5vis comparison} shows that both RIFT variants produce stable crack predictions on representative samples. They better preserve thin and low-contrast crack continuity, recover branching and connection regions, and suppress crack-like backgrounds. These observations align with the topology-aware metrics and support the sparse structural recovery perspective.

More dataset, topology-aware, and visual results are provided in \textbf{Appendix D}.

\section{Conclusion}
This paper revisits efficient crack segmentation as sparse structural recovery and proposes RIFT, a compact morphology-aligned model based on local structural preservation, cooperative directional aggregation, and lightweight multi-scale structural recovery. Experiments on four benchmarks demonstrate strong accuracy, topology preservation, transfer robustness, and deployment efficiency. Ablations show that the gains mainly arise from the structural-directional bias, while extra frequency or context enhancement brings no consistent benefit. Overall, RIFT shows that for sparse structural targets, allocating capacity to local evidence, directional continuity, and structural recovery can be more effective than complex generic feature-mixing architectures.


\bibliography{aaai2026}

\clearpage
\onecolumn

\begin{center}
{\LARGE\bfseries Rethinking Efficient Crack Segmentation with Task-Aligned Structural-Directional Modeling \\
~ \\
Supplementary Materials
\par}
\end{center}
\vspace{1em}

\input{supp}

\end{document}

%% file: supp.tex
\section{A Implementation Details of RIFT}
\label{app:implementation_details}
This appendix provides implementation details of RIFT that are not fully described in the main text. To avoid redundancy, we do not repeat the high-level motivation of the proposed framework. Instead, we focus on the architectural instantiation, block-level implementation, decoder realization, initialization strategy, and training-related settings used in the final model.

\subsection{A.1 Exact Network Instantiation}
\label{app:exact_network_instantiation}
RIFT adopts a four-stage hierarchical encoder. Given an input image, the network first applies a lightweight stem consisting of two consecutive $3\times3$ convolutional blocks, each with stride 2. The feature resolution is therefore reduced to one quarter of the input resolution before entering the encoder. Each convolutional block consists of a convolution layer, Group Normalization, and a SiLU activation. Let $C_1$ denote the channel width of the first encoder stage. The two stem layers produce $C_1/2$ and $C_1$ output channels, respectively.

The encoder contains four stages, each formed by stacking multiple Structural and Directional Modeling Blocks (SDMs). Adjacent stages are connected by a $3\times3$ convolutional block with stride 2, which performs spatial downsampling and channel expansion simultaneously. Apart from model capacity, all variants share the same structural design in the stem, encoder, multi-scale structural decoder, and prediction head.

The final model includes two variants, namely RIFT-T and RIFT-B.

\begin{table}[ht]
\centering
\caption{Architectural configurations of RIFT-T and RIFT-B.}
\label{tab:rift_variants}
\begin{tabular}{l c c c c}
\toprule
\textbf{Variant} & \textbf{Encoder widths} & \textbf{Encoder depths} & \textbf{Decoder dim} & \textbf{Directional kernel} \\
\midrule
RIFT-T & $[16, 32, 64, 128]$ & $[2, 2, 2, 1]$ & 48 & 13 \\
RIFT-B & $[32, 64, 128, 192]$ & $[2, 2, 3, 2]$ & 64 & 13 \\
\bottomrule
\end{tabular}
\end{table}

Both variants use the same directional kernel size $k=13$, the same hidden expansion ratio, and the same multi-scale decoding strategy. Their capacity difference is controlled only by the encoder width, encoder depth, and decoder dimension.

\subsection{A.2 Block-level Implementation}
\label{app:block_level_implementation}

\paragraph{Basic convolutional unit.}
The basic convolutional unit in RIFT follows a unified formulation:
\begin{equation}
\mathrm{ConvGNAct} = \mathrm{Conv2d} + \mathrm{GroupNorm} + \mathrm{SiLU}.
\end{equation}
Unless otherwise stated, convolution layers are implemented without bias. For an odd convolution kernel of size $k$, the padding is set to $(k-1)/2$ to preserve the spatial resolution. The number of groups in Group Normalization is selected adaptively rather than fixed manually. Specifically, starting from a predefined upper bound of 8, the implementation chooses the largest valid group number that exactly divides the channel dimension. This strategy ensures stable normalization across different stages and channel widths without requiring hand-crafted stage-specific settings.

\paragraph{Hidden dimension and expansion.}
For an input feature with $C$ channels, each SDM first projects it to a hidden dimension:
\begin{equation}
C_h = \max\!\bigl(C,\mathrm{round}(r\cdot C)\bigr),
\end{equation}
where the expansion ratio is fixed to $r=2.0$. This design allows local structural modeling and directional modeling to be performed in an expanded hidden space while avoiding degenerate behavior in extremely narrow configurations.

\paragraph{Structural modeling branch.}
Following the formulation in the main text, the local structural response is obtained by a depth-wise $3\times3$ convolution:
\begin{equation}
S = f_\mathrm{dw}^{3\times3}(\hat{X}),
\end{equation}
where $\hat{X}$ denotes the normalized and channel-expanded intermediate feature. This branch preserves local texture, edge evidence, and fine-grained geometric traces that are essential for crack identification.

\paragraph{Directional modeling branch.}
The directional branch consists of three depth-wise convolutions: a horizontal branch implemented by a depth-wise $1\times k$ convolution, a vertical branch implemented by a depth-wise $k\times1$ convolution, and a diagonal approximation branch implemented by a dilated depth-wise $3\times3$ convolution.

The dilation factor of the diagonal approximation branch is determined automatically from the directional kernel size:
\begin{equation}
\mathrm{dilation} = \max(1,\lfloor k/4\rfloor).
\end{equation}
Under the default setting $k=13$, this corresponds to a dilation rate of 3. The directional responses are then aggregated additively:
\begin{equation}
D = D_h + D_v + D_d.
\end{equation}
No directional competition, softmax-based directional selection, or additional directional attention branch is introduced in the final implementation.

\paragraph{Structural-directional fusion and residual output.}
The local structural response and the aggregated directional response are fused through a lightweight gating mechanism. Specifically, the gate is generated as:
\begin{equation}
G = \sigma\!\left(\mathrm{Conv}_{1\times1}(S + D)\right),
\end{equation}
where a single $1\times1$ convolution followed by a Sigmoid activation produces spatially and channel-wise adaptive modulation weights. The fused feature is then written as:
\begin{equation}
Y = S + G \odot D.
\end{equation}
Afterward, the fused representation is projected back to the original channel dimension through:
\begin{equation}
\mathrm{GroupNorm} \rightarrow \mathrm{SiLU} \rightarrow \mathrm{Conv}_{1\times1},
\end{equation}
followed by residual addition with the input feature. This realizes the SDM defined in the main text.

\paragraph{Drop path schedule.}
RIFT uses DropPath as a lightweight stochastic depth strategy at the block level. Rather than adopting a uniform drop rate, the implementation linearly increases the drop probability across all SDMs. Let $N$ denote the total number of blocks in the network and $p_{\max}$ denote the maximum drop path rate. The $i$-th block is assigned a drop probability according to a linear schedule from 0 to $p_{\max}$. In the final implementation, $p_{\max}=0.05$. This setting moderately regularizes deeper blocks while avoiding excessive disturbance to shallow feature learning.

\subsection{A.3 Multi-scale Structural Decoder and Prediction Head}
\label{app:decoder_prediction_head}

The decoder follows the multi-scale structural decoder described in the main text. Specifically, the four encoder outputs are first projected to a unified decoder width $\texttt{decoder\_dim}$ by four separate $1\times1$ ConvGNAct layers. The decoder then performs top-down reconstruction through three Multi-scale Structure Fusion (MSF) modules, progressively propagating features from $F_4$ to $F_3$, $F_2$, and $F_1$.

Each MSF module consists of two components.

\paragraph{Gate generation.}
The upsampled higher-level feature is concatenated with the current lateral feature along the channel dimension and then passed through a lightweight GateNet:
\begin{equation}
1\times1\ \mathrm{Conv}\rightarrow \mathrm{GN}\rightarrow \mathrm{SiLU}\rightarrow 1\times1\ \mathrm{Conv}\rightarrow \mathrm{Sigmoid},
\end{equation}
which produces an adaptive fusion weight.

\paragraph{Feature refinement.}
The fusion weight selectively modulates the lower-level feature, which is then added to the upsampled higher-level feature and further refined by a $3\times3$ ConvGNAct layer. This process implements the gated multi-scale fusion described by the MSF formulation in the main text.

The prediction head consists of a $3\times3$ ConvGNAct layer followed by a final $1\times1$ convolution that produces a single-channel logit map. The logits are then upsampled to the input resolution by bilinear interpolation. No auxiliary branch, boundary branch, or deep supervision is used.

\subsection{A.4 Initialization and Normalization}
\label{app:initialization_normalization}

All convolution layers are initialized with Kaiming normal initialization. When a convolution layer includes a bias term, the bias is initialized to zero. For Group Normalization, the scale parameter is initialized to 1 and the bias parameter is initialized to 0. SiLU is used consistently as the activation function throughout the network because it is empirically stable and well matched to the lightweight convolutional design.

Batch Normalization is not used in RIFT. The main reason is that crack segmentation experiments are typically conducted with small batch sizes, especially under high-resolution inputs and resource-constrained settings. Compared with Batch Normalization, Group Normalization is less sensitive to the batch size and therefore provides more stable optimization in this regime.

\subsection{A.5 Training-related Implementation Details}
\label{app:training_details}

Although the main text focuses on architectural design rather than training heuristics, we provide implementation-level training details here for reproducibility.

The supervision objective follows the setting adopted in the main text and is defined as a weighted combination of binary cross-entropy loss and Dice loss:
\begin{equation}
\mathcal{L} = \lambda_1 \mathcal{L}_{\mathrm{bce}} + \lambda_2 \mathcal{L}_{\mathrm{dice}},
\end{equation}
where the loss weights are set to:
\begin{equation}
\lambda_1 : \lambda_2 = 5 : 1.
\end{equation}
In the implementation, the binary cross-entropy term is instantiated as \texttt{BCEWithLogitsLoss}, while the Dice term is computed from $\mathrm{sigmoid}(\text{logits})$ and the binary target mask.

The default training hyperparameters are listed as follows.

\begin{table}[ht]
\centering
\caption{Default training hyperparameters.}
\label{tab:training_hyperparameters}
\begin{tabular}{l c}
\toprule
\textbf{Hyperparameter} & \textbf{Value} \\
\midrule
Epochs & 50 \\
Initial learning rate & $5\times10^{-4}$ \\
Minimum learning rate & $1\times10^{-6}$ \\
Weight decay & 0.01 \\
Training batch size & 1 \\
Validation batch size & 1 \\
\bottomrule
\end{tabular}
\end{table}

The default learning-rate scheduler is PolyLR, whose decay rule is:
\begin{equation}
\eta_t = (\eta_0-\eta_{\min})\left(1-\frac{t}{T}\right)^{0.9} + \eta_{\min},
\end{equation}
where $T$ denotes the total number of training epochs. Although the codebase also retains StepLR and CosineAnnealingWarmRestarts as optional schedulers, all main experiments reported in the paper use PolyLR.

\subsection{A.6 Complexity and Efficiency Measurement Details}
For methods evaluated in our environment, complexity and efficiency measurements are obtained under a unified implementation setting. Externally reported results marked with $\dagger$ are included only for reference. The number of parameters is computed directly from \texttt{model.parameters()}. FLOPs are estimated by registering forward hooks for \texttt{Conv2d}, \texttt{Linear}, \texttt{BatchNorm2d}, and \texttt{GroupNorm}, with convolutional FLOPs calculated from the output spatial size, kernel size, and number of groups. Peak memory consumption is recorded using \texttt{torch.cuda.max\_memory\_allocated()}. Throughput in frames per second is measured under \texttt{eval()} and \texttt{torch.inference\_mode()} with a fixed input size. A warm-up phase is performed before timing, and the final FPS is reported as the average throughput over repeated forward passes.

\subsection{A.7 Implementation Details of Ablation Variants}
Unless otherwise specified, all ablation variants share the same stem, four-stage encoder hierarchy, prediction head, and optimization objective as \textbf{RIFT-B}. Each variant changes only one targeted design factor relative to the default RIFT-B configuration, so that the corresponding performance difference can be more clearly attributed to that factor.

\textbf{Local-only} removes the entire directional crack-continuity branch and keeps only the local appearance branch in each encoder block. Specifically, the block output is constructed only from the local depth-wise $3\times3$ branch, while the directional response is set to zero. In addition, the decoder uses plain FPN-style fusion rather than gated fusion. This variant serves as the most reduced baseline and is used to test how far local low-level crack appearance alone can support segmentation without explicit crack-continuity modeling.

\textbf{Local + Gated Dec.} retains the same encoder setting as Local-only, where the directional branch is still removed and only the local appearance branch is preserved. The only difference is that the decoder is replaced by the lightweight gated decoder used in RIFT-B. This variant isolates the contribution of decoder-side gated fusion when directional crack-continuity modeling is unavailable in the encoder.

\textbf{Directional + Plain Dec.} restores the directional crack-continuity branch while keeping the decoder plain. In this variant, each encoder block uses the local branch together with the default directional aggregation:
\begin{equation}
D = D_h + D_v + D_d,
\end{equation}
where $D_h$, $D_v$, and $D_d$ denote the horizontal, vertical, and oblique-continuity responses, respectively. The last branch is implemented by a dilated depth-wise $3\times3$ convolution and serves as a lightweight approximation of oblique local continuity rather than a strict diagonal operator. However, the decoder adopts plain FPN-style fusion instead of the gated fusion used in RIFT-B. This setting verifies whether directional crack-continuity modeling remains beneficial without the gated decoder.

\textbf{RIFT-B + Freq.} extends RIFT-B by adding an explicit frequency-prior branch inside each encoder block. In the code implementation, this branch is instantiated as a fixed high-pass prior followed by a $1\times1$ projection and is injected into the gating logits:
\begin{equation}
G = \sigma\!\left(f_g(S + D) + f_{\mathrm{freq}}(\hat{X})\right).
\end{equation}
Unless otherwise noted, the default frequency prior in this variant is the Laplacian-based high-pass response. This variant is introduced to test whether explicit frequency enhancement can further strengthen thin crack responses beyond the structural cues already modeled by RIFT-B.

\textbf{RIFT-B + Context} adds a global context branch on top of RIFT-B. Specifically, a lightweight squeeze-and-excitation-style global pooling module is applied to the expanded feature $\hat{X}$, and its output is added to the gate logits together with the local and directional responses:
\begin{equation}
G = \sigma\!\left(f_g(S + D) + f_{\mathrm{ctx}}(\hat{X})\right).
\end{equation}
This variant examines whether global context information can improve crack segmentation, especially under cluttered backgrounds and large appearance variations.

\textbf{RIFT-B + Freq. + Context} simultaneously enables the explicit frequency-prior branch and the global context branch. In this variant, the gate logits are jointly modulated by local appearance, directional continuity, high-frequency response, and pooled global context. This setting corresponds to the most enriched version among the core component variants and is used to verify whether simply stacking additional enhancement cues leads to further improvement over the default RIFT-B design.

\textbf{SoftDir-Comp.} replaces the additive directional aggregation in RIFT-B with a competitive soft directional selection mechanism. Instead of directly summing the three directional responses, this variant first concatenates the horizontal, vertical, and oblique-continuity features:
\begin{equation}
Z_{\mathrm{dir}} = [D_h, D_v, D_d],
\end{equation}
then predicts a three-channel direction logit map through a $1\times1$ convolution and normalizes it with a softmax along the direction dimension:
\begin{equation}
[w_h, w_v, w_d] = \mathrm{Softmax}(f_{\mathrm{dir}}(Z_{\mathrm{dir}})).
\end{equation}
The final directional response is computed as:
\begin{equation}
D = 3 \cdot (w_h \odot D_h + w_v \odot D_v + w_d \odot D_d).
\end{equation}
In the implementation, the direction logit layer is initialized to zero so that the initial direction weights are uniform. The multiplicative factor $3$ is used to approximately match the magnitude of the original additive directional aggregation. This variant is designed to test whether competitive direction selection is preferable to the fixed additive fusion used in RIFT-B.

\textbf{Multi-Diag.} augments the default directional branch of RIFT-B with an additional dilated oblique-continuity branch at a different dilation rate. In addition to the original oblique-continuity response $D_d$, a second dilated depth-wise $3\times3$ oblique-continuity branch $D_{d2}$ is introduced. The directional aggregation becomes:
\begin{equation}
D = D_h + D_v + D_d + \tanh(\alpha)\, D_{d2},
\end{equation}
where $\alpha$ is a learnable scaling parameter initialized to zero. This initialization makes the variant equivalent to the default RIFT-B at the beginning of training and allows the second diagonal branch to be activated only when it is beneficial. This design tests whether richer oblique-continuity receptive fields can better model irregular or oblique crack patterns.

Overall, these variants are implemented as controlled modifications of RIFT-B, so that each comparison isolates a single design factor, including directional crack-continuity modeling, decoder gating, explicit frequency enhancement, global context modulation, competitive directional selection, and expanded diagonal receptive fields.

\section{B Additional Experimental Details}
\subsection{B.1 Datasets and Evaluation Protocol}
We evaluate RIFT on four public crack segmentation benchmarks: DeepCrack, CamCrack789, CrackMap, and Crack500. These datasets differ substantially in material type, scene complexity, and crack morphology, covering relatively clear thin cracks as well as low-contrast, discontinuous, branching, and mesh-like crack patterns. Except for externally reported reference results marked with $\dagger$, all reproduced experiments follow the same dataset organization, train/test splits, input resolution, and metric protocol as the MixerCSeg evaluation setting whenever applicable.

We report mIoU, ODS, OIS, and F1 as the primary evaluation metrics. Among them, mIoU measures overall region overlap, whereas ODS and OIS characterize boundary or structure detection quality under dataset-level and image-level optimal thresholds, respectively. F1 assesses the balance between precision and recall. The fixed-threshold F1 follows the adopted evaluation script and is reported for consistency with prior comparisons. Since crack segmentation performance can be sensitive to random initialization, RIFT-T and RIFT-B are evaluated with three random seeds, namely 42, 0, and 100, and both the mean and standard deviation are reported. Reproduced competing methods in the main comparison are evaluated with a single run unless otherwise specified.

\subsection{B.2 Model Variants and Training Details}
RIFT follows a unified design principle and is instantiated in two variants, RIFT-T and RIFT-B. RIFT-T is intended for lightweight deployment and efficiency evaluation, whereas RIFT-B serves as the default base model in the main quantitative comparison. Unless otherwise specified, the directional kernel size is fixed to 13.

All experiments use the same supervision objective based on a weighted combination of binary cross-entropy loss and Dice loss. Except for explicitly stated ablation settings, all reproduced methods, RIFT variants, and ablation models follow the same main evaluation setting whenever applicable. For RIFT variants and ablation models, training configurations are kept identical so that the observed performance differences mainly reflect the targeted architectural changes rather than differences in optimization strategy. The exact architectural configurations of RIFT-T and RIFT-B are provided in Appendix A.

\subsection{B.3 Hardware Environment and Efficiency Measurement}
All experiments are conducted on the same workstation: a \texttt{Dell Precision 3650 Tower} equipped with an \texttt{11th Gen Intel Core i9-11900K CPU}, \texttt{32 GB RAM}, and an \texttt{NVIDIA RTX A2000 12GB GPU}, running \texttt{Ubuntu 22.04.5 LTS (64-bit)}. Unless otherwise noted, all results produced by us, including RIFT training, ablations, topology-aware evaluation, memory statistics, and inference speed measurements, are obtained on this platform. Baseline results adopted from prior work follow the sources specified in the corresponding tables.

For resource analysis, we report FLOPs, parameter count, and peak memory consumption. For runtime efficiency, we report CUDA FPS. For methods evaluated in our environment, speed and memory are measured under the same hardware platform and input resolution. Results marked with $\dagger$ are taken from external reports at $256\times256$ and are included only as reference rather than strictly controlled comparisons.

\subsection{B.4 Topology-aware Evaluation Metrics}
\label{app:topology_metrics}

To complement conventional region-level evaluation, we further assess structural quality from three perspectives: skeleton-region consistency, centerline reconstruction, and connectivity preservation. Specifically, we report five topology-aware statistics: clDice (mean$\pm$std), Skeleton F1 (mean$\pm$std), Skeleton F1 (micro)~\cite{pantoja2022topo, shao2021mrenet}, Connectivity Ratio (mean$\pm$std), and Connectivity Ratio (global)~\cite{hyeon2025evaluating, joo2026cts}. All topology-aware metrics are computed as ratios in $[0,1]$ and multiplied by 100 when reported in tables. These metrics do not differ in the prediction target they evaluate, but in their statistical granularity: some are computed at the image level and then averaged, whereas others are aggregated over the entire dataset.

\paragraph{Preprocessing and skeleton extraction.}
Given a predicted grayscale map and a ground-truth grayscale mask, we first binarize them as:
\begin{equation}
M_p = \mathds{1}(I_p > \tau_p), \qquad
M_g = \mathds{1}(I_g > 127).
\end{equation}
In our implementation, the prediction threshold is fixed to $\tau_p=80$ for all compared methods to avoid method-specific tuning. This threshold is used only for topology-aware analysis, while the primary ODS and OIS metrics still evaluate threshold-dependent behavior. This topology threshold is independent of the fixed-threshold F1 reported by the adopted region-level evaluation script. Skeletons are then extracted from the binary masks through an iterative morphological thinning procedure based on erosion and opening residuals, using a $3\times3$ cross-shaped structuring element until all foreground pixels vanish. This yields the predicted skeleton $S_p$ and the ground-truth skeleton $S_g$. To tolerate small spatial misalignment, the skeletons are further dilated by a radius $r$ before skeleton matching and connectivity analysis. The implementation uses an elliptical structuring element, with the matching radius set to $r=5$ by default.

\paragraph{clDice (mean$\pm$std).}
clDice measures image-level skeleton-region consistency. Given the prediction mask $M_p$, ground-truth mask $M_g$, predicted skeleton $S_p$, and ground-truth skeleton $S_g$, we first compute:
\begin{equation}
P_{\mathrm{topo}} = \frac{|S_p \cap M_g|}{|S_p|}, \qquad
R_{\mathrm{topo}} = \frac{|S_g \cap M_p|}{|S_g|},
\end{equation}
which quantify the proportion of the predicted skeleton lying inside the ground-truth region and the proportion of the ground-truth skeleton covered by the predicted region, respectively. clDice is then defined as:
\begin{equation}
\mathrm{clDice} =
\frac{2P_{\mathrm{topo}}R_{\mathrm{topo}}}
{P_{\mathrm{topo}} + R_{\mathrm{topo}}}.
\end{equation}
When the denominator is zero, the score is set to zero; safe division is used whenever a pixel count in the denominator is zero. The reported clDice (mean$\pm$std) is obtained by computing clDice for each test image independently and then reporting the mean and standard deviation across the dataset.

\paragraph{Skeleton F1 (mean$\pm$std).}
Skeleton F1 quantifies image-level centerline reconstruction quality. We first dilate $S_g$ and $S_p$ by radius $r$, obtaining $\tilde{S}_g$ and $\tilde{S}_p$, respectively. We then count the number of predicted skeleton pixels that fall within $\tilde{S}_g$, denoted as $\mathrm{matched}_p$, and the number of ground-truth skeleton pixels that fall within $\tilde{S}_p$, denoted as $\mathrm{matched}_g$.

The corresponding skeleton-level precision and recall are:
\begin{equation}
P_{\mathrm{sk}} = \frac{\mathrm{matched}_p}{|S_p|}, \qquad
R_{\mathrm{sk}} = \frac{\mathrm{matched}_g}{|S_g|}.
\end{equation}
The image-level Skeleton F1 is then defined as:
\begin{equation}
F1_{\mathrm{sk}} =
\frac{2P_{\mathrm{sk}}R_{\mathrm{sk}}}
{P_{\mathrm{sk}} + R_{\mathrm{sk}}}.
\end{equation}
The reported Skeleton F1 (mean$\pm$std) is the mean and standard deviation of $F1_{\mathrm{sk}}$ over all test images.

\paragraph{Skeleton F1 (micro).}
Unlike the image-wise average above, Skeleton F1 (micro) is computed from dataset-level accumulated statistics. Specifically, we sum $\mathrm{matched}_p$, $|S_p|$, $\mathrm{matched}_g$, and $|S_g|$ over all test images, and compute the dataset-level precision and recall as:
\begin{equation}
P_{\mathrm{micro}} =
\frac{\sum_i \mathrm{matched}_{p,i}}{\sum_i |S_{p,i}|},
\qquad
R_{\mathrm{micro}} =
\frac{\sum_i \mathrm{matched}_{g,i}}{\sum_i |S_{g,i}|}.
\end{equation}
The corresponding micro-level F1 is then computed from these aggregated quantities:
\begin{equation}
F1_{\mathrm{micro}} =
\frac{2P_{\mathrm{micro}}R_{\mathrm{micro}}}
{P_{\mathrm{micro}}+R_{\mathrm{micro}}}.
\end{equation}
Unlike Skeleton F1 (mean$\pm$std), this statistic weights images according to the number of skeleton pixels they contain and therefore reflects the overall centerline reconstruction quality at the dataset scale.

\paragraph{Connectivity Ratio (mean$\pm$std).}
Connectivity Ratio evaluates the proportion of preserved crack-connected components at the image level. We first decompose the ground-truth skeleton $S_g$ into 8-connected components, each of which is regarded as one ground-truth crack segment $C_i$. Empty components are discarded.

A ground-truth segment is considered preserved only if all of the following conditions are satisfied.

\begin{enumerate}
    \item \textbf{Coverage condition.}
    The overlap between the ground-truth segment and the dilated predicted skeleton must exceed a threshold $\rho$:
    \begin{equation}
    \frac{|C_i \cap \tilde{S}_p|}{|C_i|} \ge \rho,
    \end{equation}
    where $\rho=0.7$ in the default implementation.

    \item \textbf{Local single-connectivity condition.}
    Within the local neighborhood of $C_i$, the predicted skeleton must form exactly one non-empty connected component. In practice, $C_i$ is first dilated by radius $r$ to define a local region, after which connected-component analysis is applied to the predicted skeleton inside that region.

    \item \textbf{Endpoint preservation condition.}
    If the ground-truth segment contains at least two endpoints, then all of its endpoints must be covered by the dilated predicted skeleton. Endpoints are identified as skeleton pixels with exactly one 8-neighbor.
\end{enumerate}

For a given image, if $N_{\mathrm{pres}}$ denotes the number of preserved ground-truth segments and $N_{\mathrm{tot}}$ denotes the total number of valid ground-truth segments, the image-level connectivity ratio is defined as:
\begin{equation}
\mathrm{CR} =
\frac{N_{\mathrm{pres}}}{N_{\mathrm{tot}}}.
\end{equation}
Connectivity Ratio (mean$\pm$std) is then obtained by averaging this quantity over all images and reporting the associated standard deviation.

\paragraph{Connectivity Ratio (global).}
Connectivity Ratio (global) uses dataset-level aggregation rather than image-wise averaging. Specifically, we accumulate the total number of preserved ground-truth segments and the total number of valid ground-truth segments across the full test set, and define:
\begin{equation}
CR_{\mathrm{global}} =
\frac{\sum_i N_{\mathrm{pres},i}}
{\sum_i N_{\mathrm{tot},i}}.
\end{equation}
Compared with Connectivity Ratio (mean$\pm$std), this metric assigns greater weight to samples containing more connected crack segments.

Overall, these five topology-aware statistics can be grouped into two categories. The image-level metrics, namely clDice (mean$\pm$std), Skeleton F1 (mean$\pm$std), and Connectivity Ratio (mean$\pm$std), characterize the average structural quality and its variation across individual test images. The dataset-level metrics, namely Skeleton F1 (micro) and Connectivity Ratio (global), quantify the overall centerline reconstruction quality and connectivity preservation over the full dataset. Semantically, clDice emphasizes skeleton-region consistency, Skeleton F1 emphasizes centerline reconstruction accuracy, and Connectivity Ratio directly measures whether crack connectivity is preserved.

\section{C More Ablations}
\subsection{C.1 Stage-wise Directional Kernels}
\label{app:stage_wise_directional_kernels}

\begin{table}[ht]
\centering
\caption{\textbf{Stage-wise directional kernel variants on CrackMap.} \textbf{SK-[k1,k2,k3,k4]} denotes stage-wise directional kernels, where $k_1$ to $k_4$ are used in the four encoder stages from shallow to deep. RIFT-B uses a uniform directional kernel size of $k=13$ across all stages. The best results are highlighted in bold.}
\label{tab:stage_wise_directional_kernels}
\begin{tabular}{l c c c c}
\toprule
\textbf{Method} & \textbf{mIoU} & \textbf{ODS} & \textbf{OIS} & \textbf{F1} \\
\midrule
\textbf{RIFT-B, $k=13$} & \textbf{83.0} & \textbf{80.5} & \textbf{81.3} & \textbf{80.3} \\
SK-[7,9,13,13]  & 82.9 & 80.3 & 81.2 & 78.7 \\
SK-[7,11,13,13] & 82.8 & 80.1 & 80.3 & 80.2 \\
SK-[7,11,13,17] & 82.1 & 79.4 & 80.1 & 79.8 \\
SK-[9,11,13,13] & 82.7 & 80.0 & 80.9 & 79.6 \\
\bottomrule
\end{tabular}
\end{table}

Table~\ref{tab:stage_wise_directional_kernels} analyzes whether different encoder stages require different directional kernel sizes. Although stage-wise kernels are intuitively reasonable, none of the tested configurations consistently outperforms RIFT-B with a uniform kernel size of $k=13$. SK-[7,9,13,13] achieves an mIoU close to that of RIFT-B, but its F1 score is clearly lower. In contrast, further enlarging the directional kernel in the deepest stage, as in SK-[7,11,13,17], leads to overall performance degradation. These results indicate that a uniform medium-sized directional kernel already provides a stable balance between local detail preservation and crack-continuity modeling.

\subsection{C.2 Frequency-prior Variants}
\label{app:frequency_prior_variants}

\begin{table}[ht]
\centering
\caption{\textbf{Frequency-prior variants on CrackMap.} \textbf{FP-Lap.}, \textbf{FP-Sobel}, and \textbf{FP-AvgHP} denote frequency-prior branches based on the Laplacian response, Sobel gradient magnitude, and average high-pass residual, respectively. All variants are added to RIFT-B. The best results are highlighted in bold.}
\label{tab:frequency_prior_variants}
\begin{tabular}{l c c c c}
\toprule
\textbf{Method} & \textbf{mIoU} & \textbf{ODS} & \textbf{OIS} & \textbf{F1} \\
\midrule
\textbf{RIFT-B} & \textbf{83.0} & \textbf{80.5} & \textbf{81.3} & \textbf{80.3} \\
FP-Lap.  & 81.4 & 78.1 & 78.5 & 78.3 \\
FP-Sobel & 82.2 & 79.2 & 79.7 & 79.3 \\
FP-AvgHP & 82.4 & 79.5 & 80.0 & 79.8 \\
\bottomrule
\end{tabular}
\end{table}

Table~\ref{tab:frequency_prior_variants} compares three explicit frequency-prior branches. All frequency-prior variants perform worse than RIFT-B, with FP-Lap. showing the most evident degradation. Although high-frequency responses are intuitively related to thin cracks, background textures, shadow boundaries, material joints, and other crack-like distractors can also strongly activate such responses. Therefore, directly injecting frequency-domain priors may amplify non-crack structures and weaken segmentation robustness. This observation further supports the final design choice of excluding explicit frequency-enhancement branches.

\subsection{C.3 Decoder Analysis}
\label{app:decoder_analysis}

\begin{table}[ht]
\centering
\caption{\textbf{Decoder analysis on CrackMap.} Local denotes using only the local appearance branch without directional crack modeling. Directional denotes introducing the directional crack branch. Plain Dec. denotes a standard FPN-like decoder, while Gated Dec. denotes the lightweight gated decoder used in RIFT-B. The best results are highlighted in bold.}
\label{tab:decoder_analysis}
\begin{tabular}{l c c c c}
\toprule
\textbf{Method} & \textbf{mIoU} & \textbf{ODS} & \textbf{OIS} & \textbf{F1} \\
\midrule
Local + Plain Dec.                    & 82.5 & 79.7 & 80.3 & 79.4 \\
Local + Gated Dec.                    & 82.1 & 79.0 & 79.4 & 79.3 \\
Directional + Plain Dec.              & 82.5 & 79.7 & 79.8 & 79.9 \\
\textbf{Directional + Gated Dec. (RIFT-B)} & \textbf{83.0} & \textbf{80.5} & \textbf{81.3} & \textbf{80.3} \\
\bottomrule
\end{tabular}
\end{table}

Table~\ref{tab:decoder_analysis} decouples the effects of the directional crack branch and the gated decoder. When only the local appearance branch is used, replacing the plain decoder with the gated decoder decreases mIoU from 82.5 to 82.1, suggesting that the decoder gate alone does not provide stable gains. Without directional structural cues, gated fusion may even suppress some useful fine-grained crack responses. By contrast, when the directional branch is introduced with the plain decoder, mIoU remains comparable, while F1 improves from 79.4 to 79.9. This indicates that directional continuity modeling remains beneficial even without the gated decoder.

When the directional branch is combined with the gated decoder, the model achieves the best results across all metrics. The default RIFT-B reaches 83.0 mIoU and 80.3 F1. This suggests that the gated decoder becomes more effective when guided by direction-aware structural features, enabling more reliable fusion between high-level and low-level representations. Nevertheless, this comparison also shows that the performance gain does not come from the decoder alone, but mainly from directional crack modeling in the encoder. These results further support the design division in RIFT: the encoder models irregular crack continuity, while the decoder serves as a lightweight structural recovery module rather than a complex semantic enhancement module.

Overall, the gated decoder is useful only when combined with directional crack features, whereas directional continuity modeling remains the key source of the performance improvement of RIFT.

\section{Additional Experimental Results}
\label{app:additional_results}

\subsection{D.1 Additional Results of Topology-aware Evaluation}
\label{app:additional_topology_results}

\begin{table}[ht]
\centering
\small
\setlength\tabcolsep{0.7mm}
\caption{\textbf{Image-level topology-aware evaluation on four crack segmentation benchmarks (\%).} Comparison of representative methods in terms of clDice (mean$\pm$std), Skeleton F1 (mean$\pm$std), and Connectivity Ratio (CR, mean$\pm$std) on Crack500, CrackMap, DeepCrack, and CamCrack789. clDice measures image-level skeleton-region consistency, Skeleton F1 measures image-level centerline reconstruction quality, and CR measures the image-level proportion of preserved crack-connected segments. The symbol ``-'' indicates that the corresponding connectivity statistic is unavailable under the current evaluation setting.}
\label{tab:image_level_topology}
\begin{tabular}{l c c c c c c c c c c c c}
\toprule
\multirow{2}{*}{\textbf{Method}} 
& \multicolumn{3}{c}{\textbf{Crack500}} 
& \multicolumn{3}{c}{\textbf{CrackMap}} 
& \multicolumn{3}{c}{\textbf{DeepCrack}} 
& \multicolumn{3}{c}{\textbf{CamCrack789}} \\
\cmidrule(lr){2-4}
\cmidrule(lr){5-7}
\cmidrule(lr){8-10}
\cmidrule(lr){11-13}
& \textbf{clDice} & \textbf{Skel. F1} & \textbf{CR}
& \textbf{clDice} & \textbf{Skel. F1} & \textbf{CR}
& \textbf{clDice} & \textbf{Skel. F1} & \textbf{CR}
& \textbf{clDice} & \textbf{Skel. F1} & \textbf{CR} \\
\midrule
SCSegamba & 77.9$\pm$17.8 & 63.2$\pm$20.1 & 6.5$\pm$17.3 
& 81.9$\pm$5.4 & 83.8$\pm$5.5 & - 
& 94.9$\pm$3.9 & 93.4$\pm$6.4 & 13.4$\pm$28.8 
& 89.0$\pm$8.9 & 91.6$\pm$8.1 & 11.5$\pm$28.6 \\
MixerCSeg & 80.5$\pm$17.8 & 65.2$\pm$20.5 & 4.7$\pm$13.5 
& 89.7$\pm$5.2 & 86.9$\pm$6.6 & - 
& 95.5$\pm$3.4 & 93.6$\pm$6.6 & 13.1$\pm$28.9 
& 90.8$\pm$9.0 & 93.3$\pm$7.6 & 10.7$\pm$27.4 \\
RIFT-T & 82.4$\pm$17.0 & 67.4$\pm$20.8 & 5.8$\pm$16.5 
& 89.5$\pm$4.6 & 86.1$\pm$6.6 & - 
& 96.2$\pm$3.2 & 94.7$\pm$4.9 & 14.9$\pm$30.2 
& 92.2$\pm$7.1 & 94.5$\pm$6.4 & 16.1$\pm$33.3 \\
RIFT-B & 82.0$\pm$16.7 & \textbf{68.3$\pm$20.4} & 6.2$\pm$16.8 
& \textbf{91.1$\pm$4.3} & \textbf{88.0$\pm$5.9} & - 
& \textbf{96.3$\pm$2.9} & \textbf{94.8$\pm$5.0} & \textbf{15.9$\pm$31.4} 
& \textbf{93.0$\pm$6.5} & \textbf{94.7$\pm$6.5} & 13.4$\pm$30.9 \\
\bottomrule
\end{tabular}
\end{table}

Table~\ref{tab:image_level_topology} supplements the image-level topology-aware evaluation, including clDice (mean$\pm$std), Skeleton F1 (mean$\pm$std), and Connectivity Ratio (mean$\pm$std). Overall, RIFT maintains stable advantages in skeleton-level structural recovery. Specifically, RIFT-B achieves the highest Skeleton F1 (mean$\pm$std) on Crack500, CrackMap, DeepCrack, and CamCrack789, indicating the most stable image-level centerline reconstruction quality. RIFT-T also outperforms the competing methods on most datasets, suggesting that this structural advantage does not depend solely on larger model capacity.

For clDice (mean$\pm$std), RIFT also achieves leading results on multiple datasets. The improvements are especially evident on DeepCrack and CamCrack789, where both RIFT-T and RIFT-B outperform the competing methods. On CrackMap, RIFT-B also obtains the best result. These results show that RIFT improves not only centerline reconstruction quality, but also skeleton-region consistency. In contrast, Connectivity Ratio (mean$\pm$std) has much lower absolute values, indicating that this metric imposes a stricter criterion on structural connectivity. Even under this strict setting, RIFT still shows better or competitive connectivity preservation on the evaluable datasets, particularly on DeepCrack and CamCrack789.

These three metrics characterize different structural properties. clDice emphasizes the coverage consistency between crack regions and skeletons, Skeleton F1 focuses on centerline localization quality, and Connectivity Ratio directly measures whether connected crack segments are completely preserved. Therefore, Table~\ref{tab:image_level_topology} further demonstrates that the advantage of RIFT is not limited to region-level segmentation, but extends to finer-grained skeleton recovery and structural preservation.

\begin{table}[ht]
\centering
\caption{\textbf{Dataset-level topology-aware statistics on four crack segmentation benchmarks (\%).} Comparison of representative methods using Skeleton F1 (micro) and Connectivity Ratio (CR, global) on Crack500, CrackMap, DeepCrack, and CamCrack789. Unlike Table~\ref{tab:image_level_topology}, these metrics are not computed by averaging image-level scores. Instead, Skeleton F1 (micro) is computed from accumulated skeleton matches over the full dataset, while CR (global) is computed from accumulated preserved and total connected crack segments over the full dataset. The symbol ``-'' indicates that the corresponding connectivity statistic is unavailable under the current evaluation setting.}
\label{tab:dataset_level_topology}
\begin{tabular}{l c c c c c c c c}
\toprule
\multirow{2}{*}{\textbf{Method}} 
& \multicolumn{2}{c}{\textbf{Crack500}} 
& \multicolumn{2}{c}{\textbf{CrackMap}} 
& \multicolumn{2}{c}{\textbf{DeepCrack}} 
& \multicolumn{2}{c}{\textbf{CamCrack789}} \\
\cmidrule(lr){2-3}
\cmidrule(lr){4-5}
\cmidrule(lr){6-7}
\cmidrule(lr){8-9}
& \textbf{Skel. F1} & \textbf{CR}
& \textbf{Skel. F1} & \textbf{CR}
& \textbf{Skel. F1} & \textbf{CR}
& \textbf{Skel. F1} & \textbf{CR} \\
\midrule
SCSegamba & 61.7 & \textbf{10.7} & 83.4 & - & 93.1 & 24.2 & 90.7 & 12.1 \\
MixerCSeg & 65.0 & 9.2 & 86.1 & - & 92.9 & 20.4 & 92.3 & 13.8 \\
RIFT-T    & 66.8 & 10.4 & 85.3 & - & 94.4 & \textbf{27.4} & 93.6 & \textbf{18.2} \\
RIFT-B    & \textbf{67.8} & 10.6 & \textbf{87.4} & - & \textbf{94.6} & 26.9 & \textbf{93.7} & 16.0 \\
\bottomrule
\end{tabular}
\end{table}

Table~\ref{tab:dataset_level_topology} further reports topology-aware results based on dataset-level aggregated statistics, including Skeleton F1 (micro) and Connectivity Ratio (global). Different from the image-wise averages in Table~\ref{tab:image_level_topology}, these two metrics evaluate structural recovery from accumulated skeleton-pixel matches and accumulated preserved connected segments over the entire test set. Therefore, they better reflect structural recovery capability at the dataset scale.

In terms of Skeleton F1 (micro), RIFT achieves the best or near-best performance on all four datasets. RIFT-B obtains the highest results on Crack500, CrackMap, DeepCrack, and CamCrack789. This indicates that RIFT can more reliably recover crack centerline structures under both image-level averaging and dataset-level aggregation. For Connectivity Ratio (global), RIFT also performs better on most evaluable datasets. RIFT-T achieves the highest values on DeepCrack and CamCrack789, while RIFT-B is slightly better than RIFT-T on Crack500. These results suggest that the advantage of RIFT is reflected not only in skeleton-pixel matching, but also in the preservation of connected crack segments.

Taken together, Tables~\ref{tab:image_level_topology} and~\ref{tab:dataset_level_topology} show that the structural advantage of RIFT is consistent under two statistical granularities. On the one hand, RIFT achieves higher average skeleton recovery quality at the image level. On the other hand, it more effectively preserves crack centerlines and connected structures at the whole-dataset level. This finding is consistent with the design goal of RIFT, which aims to recover irregular crack structures more robustly through local appearance modeling and directional continuity aggregation, rather than merely improving surface-level region overlap.

\subsection{D.2 Crack Segmentation under Complex Backgrounds}
\label{app:complex_backgrounds}

To further evaluate model robustness under complex background interference, we conduct additional experiments on the TUT dataset.

\noindent\textbf{TUT dataset.}
TUT is a recently introduced structural crack segmentation benchmark designed to evaluate crack segmentation under diverse material appearances and complex background interference~\cite{liu2024staircase}. It contains 1,408 RGB images with pixel-level annotations, including 987 training images, 139 validation images, and 282 test images. Different from conventional road- or pavement-centered crack datasets, TUT covers eight scenarios, including bitumen, cement, bricks, plastic runways, tiles, metal materials, generator blades, and underground pipelines. Such diverse scenes introduce considerable variations in crack morphology, surface texture, illumination, and background distractors, making TUT a challenging benchmark for assessing model robustness beyond standard crack segmentation settings. In this work, we use TUT as an additional evaluation dataset to examine whether RIFT can maintain reliable structure recovery under stronger texture interference and cross-surface appearance variations.

The quantitative results are reported in Table~\ref{tab:tut_region}. All methods in this TUT evaluation are trained and evaluated by us under the same data split, input resolution, loss setting, and metric protocol, unless otherwise specified. Overall, SCSegamba achieves the best region-level and boundary-level results on this dataset, reaching 84.8 mIoU, 82.0 ODS, 82.6 OIS, and 83.9 F1. Nevertheless, RIFT-B obtains stable second-best results across all four metrics and clearly outperforms MixerCSeg, indicating that RIFT can still maintain strong segmentation performance under complex backgrounds, texture interference, and non-ideal imaging conditions.

Comparing the two RIFT variants, RIFT-B outperforms RIFT-T in mIoU, ODS, OIS, and F1, which is consistent with its higher model capacity. This also suggests that moderately increasing capacity is helpful for improving region overlap and overall detection quality under complex background conditions. Meanwhile, the gap between RIFT-B and SCSegamba remains relatively small, showing that the proposed compact structural prior remains competitive without relying on complex heterogeneous feature mixing. Instead, it achieves stable and competitive performance under strong background interference.

\begin{table}[ht]
\centering
\caption{\textbf{Quantitative comparison on the TUT dataset under complex background conditions (\%).} Comparison of representative methods on the TUT dataset using mIoU, ODS, OIS, and F1. The best and second-best results for each metric are highlighted in bold and underlined, respectively.}
\label{tab:tut_region}
\begin{tabular}{l c c c c}
\toprule
\textbf{Method} & \textbf{mIoU} & \textbf{ODS} & \textbf{OIS} & \textbf{F1} \\
\midrule
MixerCSeg & 81.7 & 77.6 & 78.8 & 79.9 \\
SCSegamba & \textbf{84.8} & \textbf{82.0} & \textbf{82.6} & \textbf{83.9} \\
RIFT-T & 83.9 & 80.8 & 81.4 & 82.6 \\
RIFT-B & \underline{84.4} & \underline{81.5} & \underline{82.2} & \underline{83.5} \\
\bottomrule
\end{tabular}
\end{table}

\begin{table}[ht]
\centering
\caption{\textbf{Topology-aware evaluation on the TUT dataset under complex background conditions (\%).} Comparison of representative methods on the TUT dataset using topology-aware metrics, including clDice (mean$\pm$std), Skeleton F1 (mean$\pm$std), Connectivity Ratio (CR, mean$\pm$std), Skeleton F1 (micro), and Connectivity Ratio (CR, global). clDice measures image-level skeleton-region consistency, Skeleton F1 measures centerline reconstruction quality, and CR measures crack connectivity preservation. The best and second-best results are highlighted in bold and underlined, respectively.}
\label{tab:tut_topology}
\begin{tabular}{l c c c c c}
\toprule
\textbf{Method} & \textbf{clDice} & \textbf{Skel. F1} & \textbf{CR} & \textbf{Skel. F1 (micro)} & \textbf{CR (global)} \\
\midrule
MixerCSeg & 87.4$\pm$11.3 & 89.8$\pm$10.7 & 22.7$\pm$34.5 & 89.9 & 26.9 \\
SCSegamba & \textbf{91.3$\pm$9.1} & \textbf{93.2$\pm$8.6} & 27.1$\pm$36.9 & \textbf{93.6} & 30.1 \\
RIFT-T & \underline{90.6$\pm$9.4} & \underline{92.4$\pm$9.1} & \underline{35.2$\pm$40.9} & 92.7 & \underline{36.4} \\
RIFT-B & 90.8$\pm$9.7 & 92.5$\pm$9.7 & \textbf{36.1$\pm$40.7} & \underline{93.0} & \textbf{40.2} \\
\bottomrule
\end{tabular}
\end{table}

Table~\ref{tab:tut_topology} further compares the structural recovery capability of different methods on the TUT dataset from a topology-aware perspective. Unlike the region-level results in Table~\ref{tab:tut_region}, these metrics focus more on crack centerlines, skeleton-region consistency, and connected-structure preservation. SCSegamba achieves the best results in clDice (mean$\pm$std), Skeleton F1 (mean$\pm$std), and Skeleton F1 (micro), suggesting stronger skeleton-region consistency and centerline reconstruction. In contrast, RIFT-B obtains the highest Connectivity Ratio (mean$\pm$std) and Connectivity Ratio (global), reaching 36.1$\pm$40.7 and 40.2, respectively. This indicates that RIFT-B is more favorable for preserving the completeness of connected crack segments under complex backgrounds.

These results reveal different emphases in structural recovery among the compared methods. SCSegamba is stronger in skeleton matching and centerline reconstruction, whereas RIFT-B shows a clearer advantage in connectivity preservation. This is consistent with the design goal of RIFT: the model emphasizes maintaining global crack connectivity through local appearance modeling and directional continuity aggregation, rather than merely pursuing local overlap consistency between skeletons and regions. Meanwhile, RIFT-T also outperforms MixerCSeg on several topology-aware metrics, further demonstrating that the structural prior of RIFT remains effective for crack segmentation under complex background interference even in the lightweight configuration.

\subsection{D.3 Cross-dataset Transfer Evaluation}
\label{app:cross_dataset_transfer}

\begin{table}[ht]
\centering
\small
\caption{\textbf{Cross-dataset transfer evaluation on four crack segmentation benchmarks (mIoU, \%).} Each method is trained on one source dataset and evaluated on the remaining target datasets. The final column reports the average transfer performance across all source-target pairs. The best result in each column is highlighted in bold.}
\label{tab:cross_dataset_transfer}
\begin{tabular}{l c c c c c c c c c c c c c}
\toprule
\multirow{2}{*}{\textbf{Method}}
& \multicolumn{3}{c}{\textbf{DeepCrack(DeepC)}} 
& \multicolumn{3}{c}{\textbf{Crack500(C500)}} 
& \multicolumn{3}{c}{\textbf{CamCrack789(C789)}} 
& \multicolumn{3}{c}{\textbf{CrackMap(CMap)}} 
& \multirow{2}{*}{\textbf{Avg.}} \\
\cmidrule(lr){2-4}
\cmidrule(lr){5-7}
\cmidrule(lr){8-10}
\cmidrule(lr){11-13}
& C500 & C789 & CMap
& DeepC & C789 & CMap
& DeepC & C500 & CMap
& DeepC & C789 & C500
& \\
\midrule
CarNet    & 51.5 & 75.6 & 57.5 & 78.5 & 65.5 & 72.2 & 84.6 & 53.4 & 62.4 & 65.2 & 56.2 & 55.5 & 64.8 \\
RINDNet   & 54.9 & 73.6 & 60.2 & 81.7 & 68.7 & 75.3 & 82.5 & 56.8 & 64.3 & 64.2 & 55.3 & 59.4 & 66.4 \\
DTrCNet   & 56.4 & 74.9 & 63.6 & 83.4 & 70.5 & 78.0 & 84.3 & 58.6 & 67.7 & 70.2 & 60.3 & 62.9 & 69.2 \\
SCSegamba & 57.9 & 78.9 & 68.5 & 85.4 & 73.8 & 79.6 & 89.2 & 63.0 & 74.2 & \textbf{83.1} & 70.4 & \textbf{72.0} & 74.6 \\
MixerCSeg & 63.5 & 78.2 & 60.5 & 86.3 & 72.2 & 81.3 & 90.5 & 61.1 & 71.1 & 68.0 & 58.5 & 60.6 & 71.0 \\
RIFT-T    & 64.4 & \textbf{81.7} & \textbf{73.8} & \textbf{89.1} & \textbf{77.9} & \textbf{82.9} & 90.8 & \textbf{64.7} & \textbf{77.8} & 82.7 & 69.6 & 68.3 & \textbf{77.0} \\
RIFT-B    & \textbf{65.3} & 81.3 & 68.8 & \textbf{89.1} & 76.7 & 81.5 & \textbf{91.3} & 64.2 & 77.1 & 83.0 & \textbf{70.8} & 69.9 & 76.6 \\
\bottomrule
\end{tabular}
\end{table}

Table~\ref{tab:cross_dataset_transfer} reports cross-dataset transfer results among four crack segmentation benchmarks. The results show that RIFT achieves the strongest overall transfer performance. In particular, RIFT-T obtains the best average result of 77.0, while RIFT-B follows closely with an average score of 76.6. Compared with previous methods, RIFT improves transfer robustness across most source-target combinations, indicating that the proposed local structural modeling and directional continuity aggregation provide more transferable crack representations.

A closer inspection shows that RIFT-T performs particularly well in several challenging transfer settings, such as DeepCrack to CamCrack789, DeepCrack to CrackMap, Crack500 to CamCrack789, Crack500 to CrackMap, CamCrack789 to Crack500, and CamCrack789 to CrackMap. RIFT-B achieves the best results in other settings, including DeepCrack to Crack500, CamCrack789 to DeepCrack, and CrackMap to CamCrack789. These results suggest that the two variants exhibit slightly different generalization behaviors: RIFT-B benefits from stronger capacity in some target domains, while RIFT-T may provide better regularization and transfer stability in others.

Overall, the cross-dataset results support the robustness of RIFT beyond within-dataset evaluation. Since crack appearance, background texture, imaging quality, and annotation characteristics can vary substantially across datasets, the consistently high transfer performance indicates that RIFT does not merely overfit dataset-specific visual patterns. Instead, it captures structural crack cues that are more stable across domains.

\subsection{D.4 Training Loss}
\label{app:training_loss}

\begin{figure}[ht]
\centering
\includegraphics[width=0.55\linewidth]{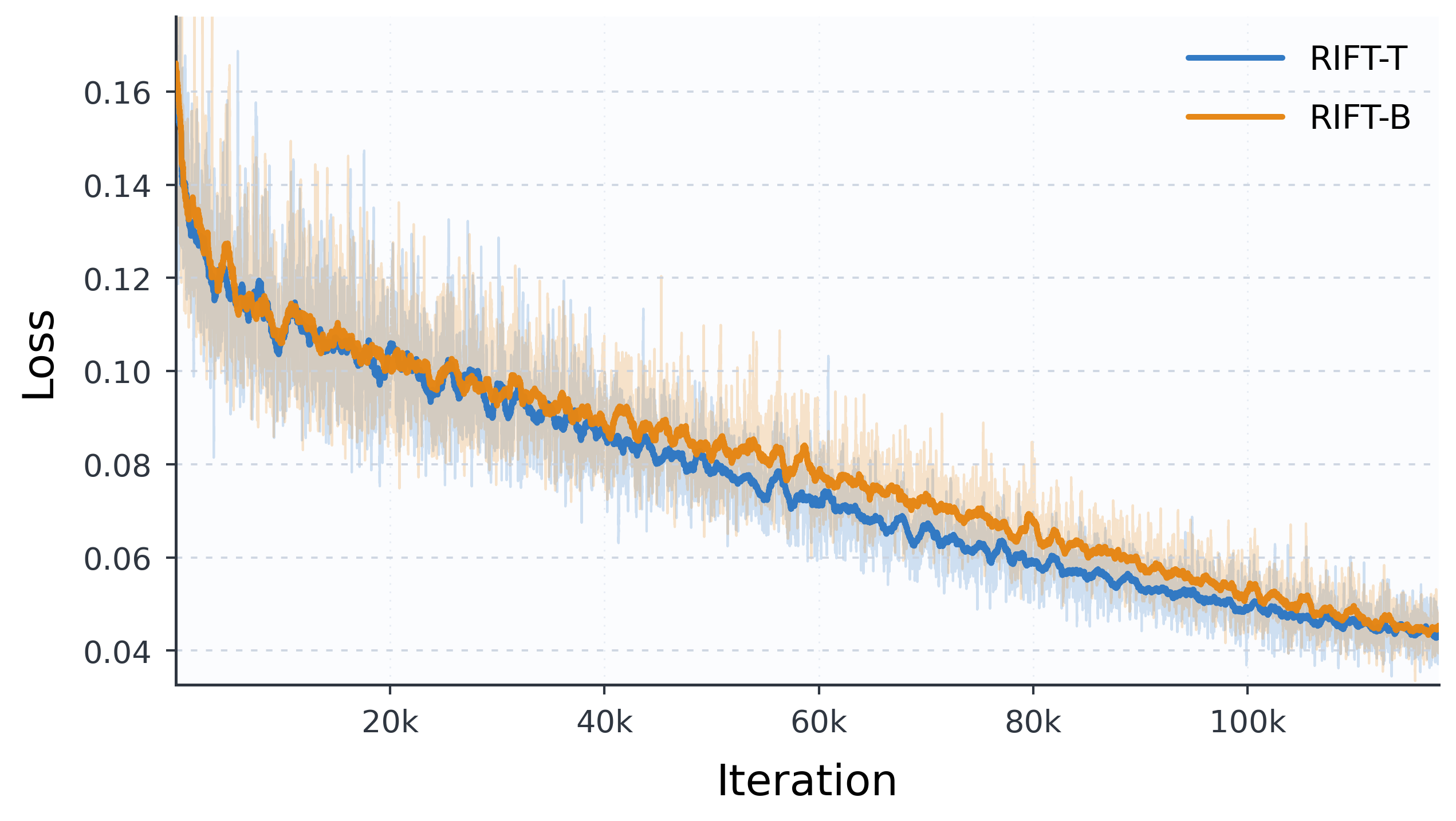}
\caption{\textbf{Training loss curves of RIFT-T and RIFT-B.} Training loss trajectories of RIFT-T and RIFT-B over iterations on the Crack500 dataset. The solid lines denote the smoothed loss curves, while the faint lines indicate the raw mini-batch loss values. Both variants exhibit stable optimization behavior and consistent convergence throughout training. RIFT-B shows a slightly higher loss in the middle stage of training, whereas the two models converge to comparable final loss levels.}
\label{fig:training_loss}
\end{figure}

Figure~\ref{fig:training_loss} shows the training loss trajectories of RIFT-T and RIFT-B. Overall, both variants exhibit stable optimization behavior. As the number of iterations increases, the training loss continuously decreases, with no obvious uncontrolled oscillation or late-stage divergence. This indicates that the structural design of RIFT can be optimized stably and has favorable training behavior.

More specifically, both RIFT-T and RIFT-B show a rapid loss decrease in the early training stage, followed by a smoother convergence process. Compared with RIFT-T, RIFT-B shows a slightly higher loss in the middle stage of training, which may reflect different optimization dynamics caused by its larger capacity. As training proceeds, however, the losses of the two models gradually become close and eventually converge to comparable final levels. This suggests that although RIFT-B has larger capacity, its optimization process does not become unstable due to increased model complexity. Instead, it supports stronger final performance while maintaining stable convergence.

In addition, the raw mini-batch loss curves show a certain degree of fluctuation in the early training stage for both variants, which is common in crack segmentation with small-batch training. Such fluctuations gradually decrease as training progresses, further indicating that the model parameters enter a more stable optimization region. Overall, Figure~\ref{fig:training_loss} demonstrates that both RIFT-T and RIFT-B have stable optimization behavior, while the larger RIFT-B variant maintains stable convergence despite its increased capacity. This observation is consistent with the quantitative results reported in the main paper.

\subsection{D.5 More Visual Results}
This section provides additional visual evidence to complement the feature visualization and qualitative comparison in the main paper. The stage-wise feature visualizations follow the same format as Fig.~4, while the qualitative comparisons follow the same format as Fig.~5. Compared with the main-paper figures, the additional qualitative results are shown without manually annotated highlight boxes, so as to present the overall segmentation behavior in a less guided manner.

\begin{figure}[ht]
    \centering
    \includegraphics[width=1\linewidth]{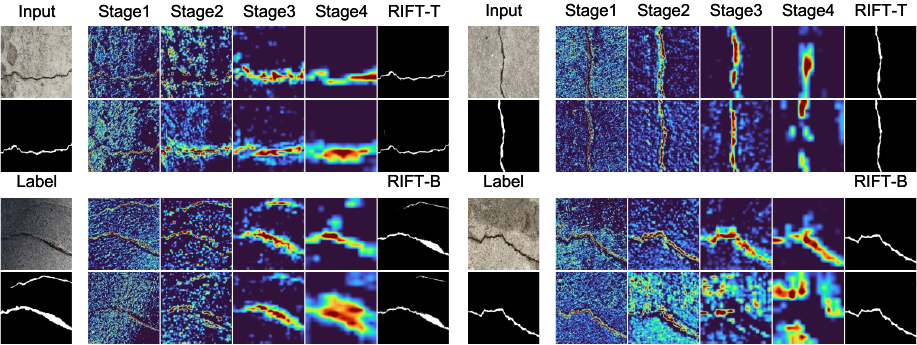}
    \caption{Additional stage-wise feature visualization (Part 1/2).}
    \label{fig:feat1}
\end{figure}

\begin{figure}[ht]
    \centering
    \includegraphics[width=1\linewidth]{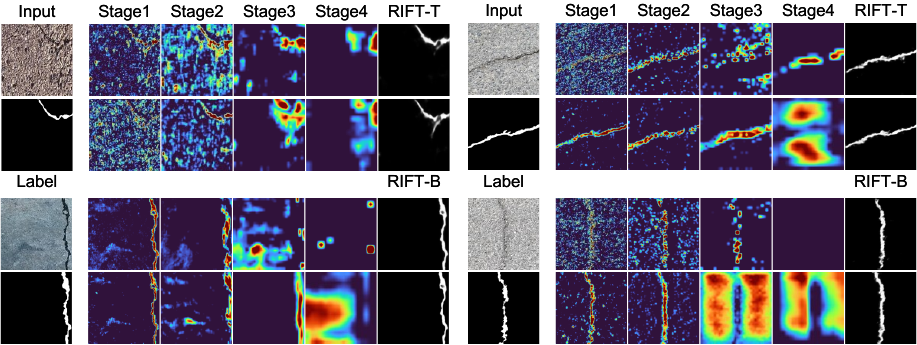}
    \caption{Additional stage-wise feature visualization (Part 2/2).}
    \label{fig:feat2}
\end{figure}

Figures~\ref{fig:feat1} and~\ref{fig:feat2} show additional stage-wise feature visualizations of RIFT-T and RIFT-B. These examples further illustrate how the learned representations evolve from local low-level responses to more structure-aware crack activations across stages. In general, shallow stages preserve fine local details and texture-sensitive responses, whereas deeper stages tend to shift from scattered local texture responses toward more elongated crack-related structures. These feature responses are used for representation analysis and should not be interpreted as final predictions. The comparison between RIFT-T and RIFT-B also suggests that the base variant produces more stable and spatially concentrated responses in challenging regions, which is consistent with its stronger quantitative performance.

\begin{figure}[ht]
    \centering
    \includegraphics[width=1\linewidth]{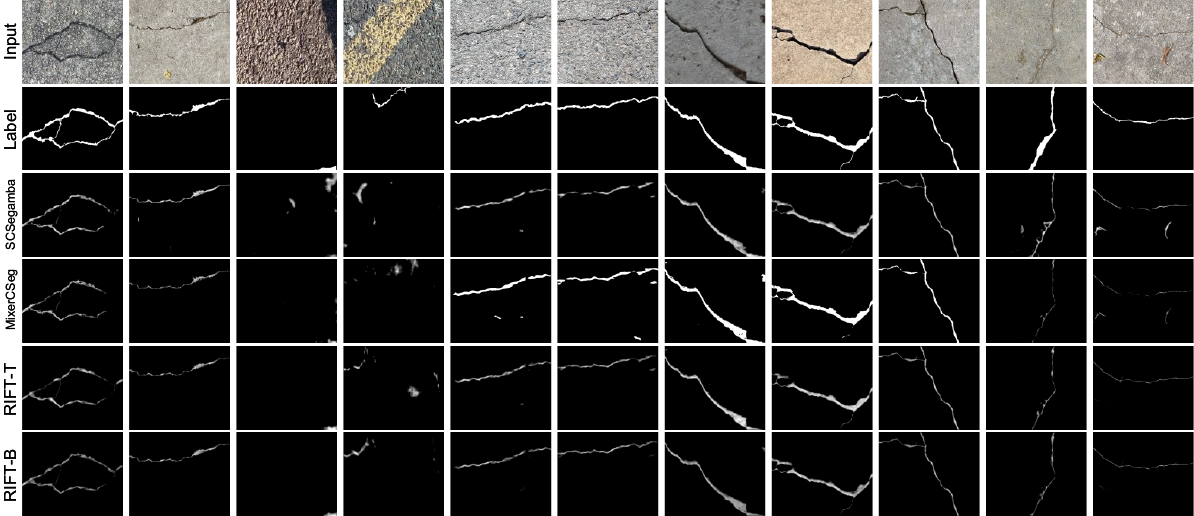}
    \caption{Additional qualitative comparison (Part 1/6). These examples are shown without manually annotated boxes to provide an unguided view of the overall segmentation behavior.}
    \label{fig:vis1}
\end{figure}

\begin{figure}[ht]
    \centering
    \includegraphics[width=1\linewidth]{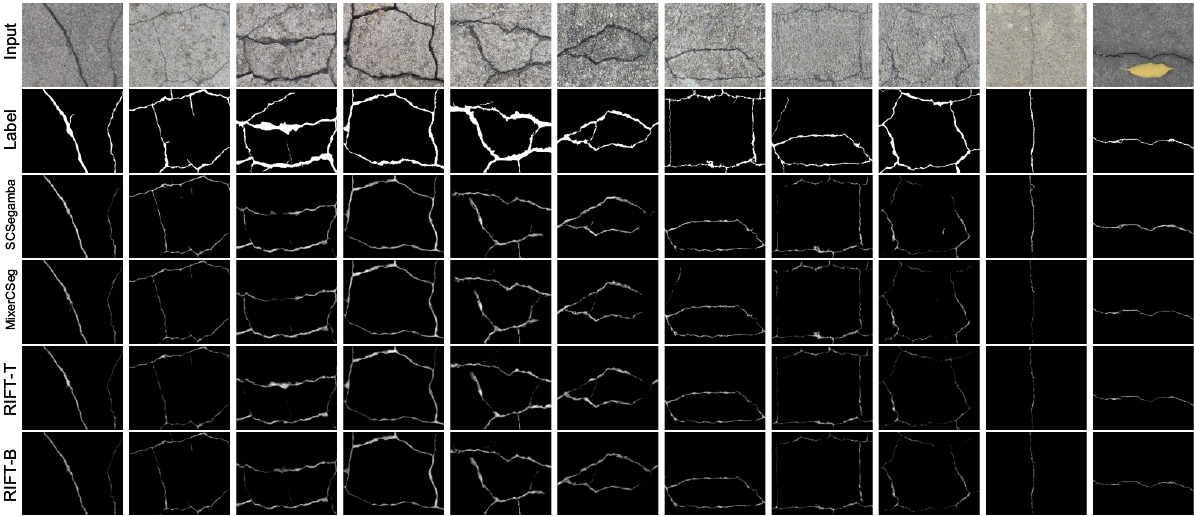}
    \caption{Additional qualitative comparison (Part 2/6).}
    \label{fig:vis2}
\end{figure}

\begin{figure}[ht]
    \centering
    \includegraphics[width=1\linewidth]{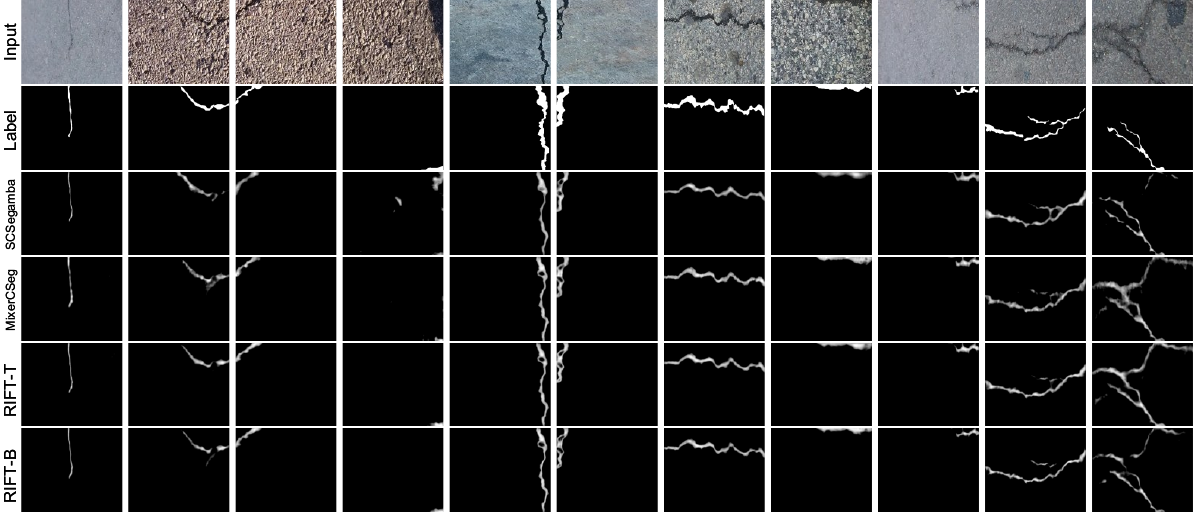}
    \caption{Additional qualitative comparison (Part 3/6).}
    \label{fig:vis3}
\end{figure}

\begin{figure}[ht]
    \centering
    \includegraphics[width=1\linewidth]{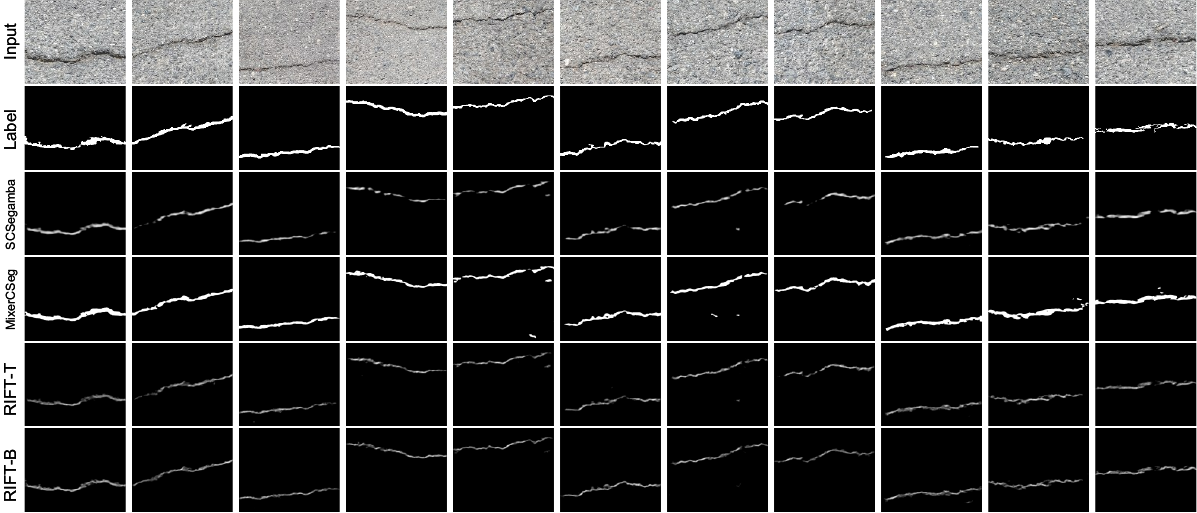}
    \caption{Additional qualitative comparison (Part 4/6).}
    \label{fig:vis4}
\end{figure}

\begin{figure}[ht]
    \centering
    \includegraphics[width=1\linewidth]{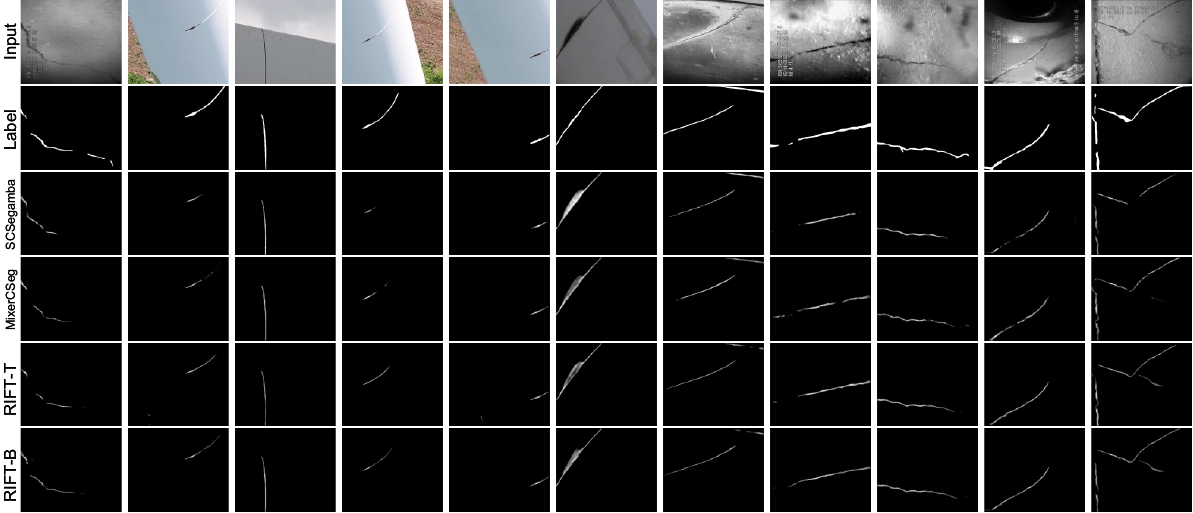}
    \caption{Additional qualitative comparison (Part 5/6).}
    \label{fig:vis5}
\end{figure}

\begin{figure}[ht]
    \centering
    \includegraphics[width=1\linewidth]{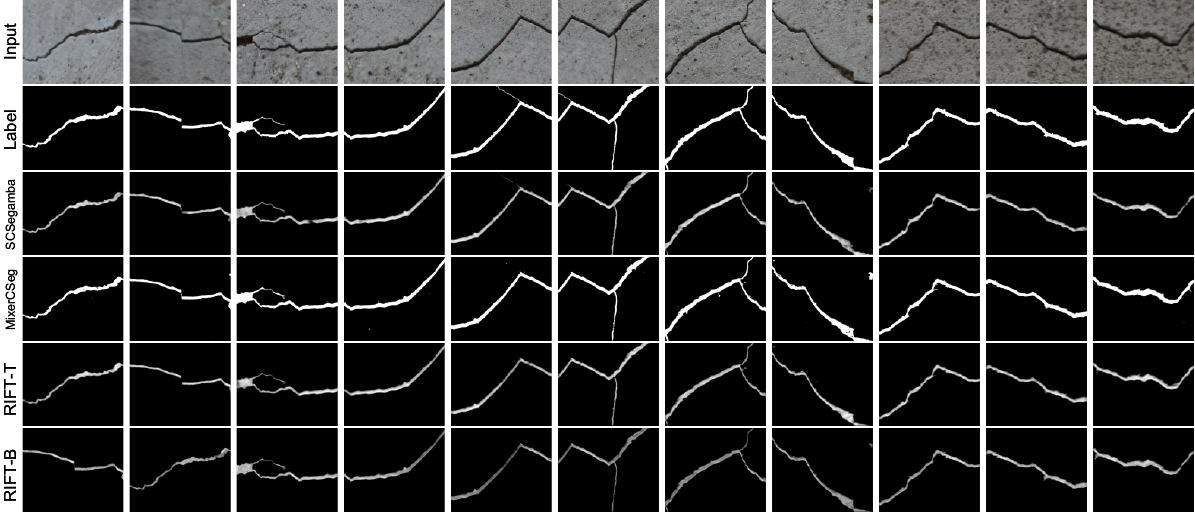}
    \caption{Additional qualitative comparison (Part 6/6).}
    \label{fig:vis6}
\end{figure}

Figures~\ref{fig:vis1}--\ref{fig:vis6} provide additional qualitative comparisons on diverse crack images. These cases include thin cracks, discontinuous cracks, low-contrast regions, and backgrounds with texture or shadow interference. The additional results show that RIFT generally preserves crack continuity in challenging cases while reducing some responses to non-crack textures. In particular, the predictions tend to recover elongated and fragmented crack structures more coherently in representative cases, which supports the motivation of jointly modeling local structural evidence and directional continuity.

